\documentclass{article}

\usepackage[preprint]{neurips_2026}
\usepackage[table]{xcolor}

\usepackage[utf8]{inputenc} 
\usepackage[T1]{fontenc}    
\usepackage{url}            
\usepackage{booktabs}       
\usepackage{amsfonts}       
\usepackage{nicefrac}       
\usepackage{microtype}      
\usepackage{xcolor}         
\usepackage{amsmath}
\usepackage{wrapfig}

\definecolor{mydarkblue}{HTML}{3F3FFF}
\definecolor{mydarkgreen}{HTML}{228B22}
\usepackage[colorlinks,
linkcolor=mydarkblue,
citecolor=mydarkblue]{hyperref}

\usepackage{graphicx} 
\usepackage{makecell}
\usepackage[table]{xcolor}
\usepackage{booktabs}
\usepackage{makecell}
\usepackage{multirow}
\usepackage{algorithm}
\usepackage{algpseudocode}
\usepackage{graphicx}
\usepackage{subcaption}
\usepackage{pifont}

\usepackage[most]{tcolorbox}
\usepackage[table]{xcolor}
\usepackage{wrapfig}
\usepackage{float}

\usepackage{minitoc}

\title{Detecting Distillation Data from Reasoning Models}

\author{
Hengxiang Zhang\textsuperscript{1},\enspace
Hyeong Kyu Choi\textsuperscript{2},\enspace
Sharon Li\textsuperscript{2},\enspace
Hongxin Wei \textsuperscript{1}\thanks{Corresponding author (\texttt{weihx@sustech.edu.cn})} \\
\textsuperscript{1}Department of Statistics and Data Science, Southern University of Science and Technology \\
\textsuperscript{2} Department of Computer Sciences,
University of Wisconsin–Madison\\
}

\newtcolorbox[auto counter, number within=section]{promptbox}[2][]{
    float,
    breakable,
    title={Example~\thetcbcounter: #2},
    label={#1}
}

\begin{document}
\doparttoc 
\faketableofcontents 

\maketitle

\begin{abstract}
Reasoning distillation has emerged as a prevailing paradigm for transferring reasoning capabilities from large reasoning models to small language models.
Yet, reasoning distillation risks data contamination: benchmark data may inadvertently be included in the distillation data, thereby inflating model performance metrics.
In this work, we formally define the distillation data detection task, which determines whether a given question is included in the model’s distillation data.
The unique challenge of this task lies in the partial availability of distillation data.
To address this, we propose \textit{Token Probability Deviation} (\textbf{TPD}), a detection method that leverages the probability patterns of \textit{output tokens} generated by the model instead of \textit{input tokens}.
Our method is motivated by the observation that seen questions tend to elicit more near-deterministic tokens generated by the models than unseen ones.
Our TPD score is thus designed to quantify the token-level deviation of generated tokens from a high-confidence reference probability.
Consequently, seen questions can yield substantially lower TPD scores than unseen ones, enabling strong detection performance.
Extensive experiments demonstrate the effectiveness of our approach, 
improving detection AUC by up to 31\% on distillation datasets.


\end{abstract}

\section{Introduction}
Large Reasoning Models (LRMs) have shown impressive performance on complex tasks like mathematical reasoning and coding problems~\citep{jaech2024openai, guo2025deepseek, yang2025qwen3, xai2025grok3}. 
By articulating intermediate steps via Chain-of-Thought (CoT), LRMs dynamically allocate extra compute to challenging problems. 
However, such reasoning capabilities are typically limited to LRMs with extremely large parameters, hindering practical deployment in resource-constrained settings~\citep{wei2022chain}. To address this, recent studies have explored reasoning distillation, transferring reasoning abilities from LRMs to Small Language Models (SLMs) by simulating reasoning traces~\citep{chen2025distilling, ye2025limo, muennighoffs1, liu2025qfft}. This paradigm has been widely applied in cutting-edge models, such as DeepSeek-R1 series~\citep{guo2025deepseek}, Sky-T1-32B-preview~\citep{sky_t1_2025}, Bespoke-32B~\citep{bespoke_stratos}, and Qwen3 series~\citep{yang2025qwen3}.

\begin{figure*}[t]
    \vspace{-7mm}
    \centering
    \includegraphics[width=0.9\textwidth, height=0.27\textheight, keepaspectratio=False]{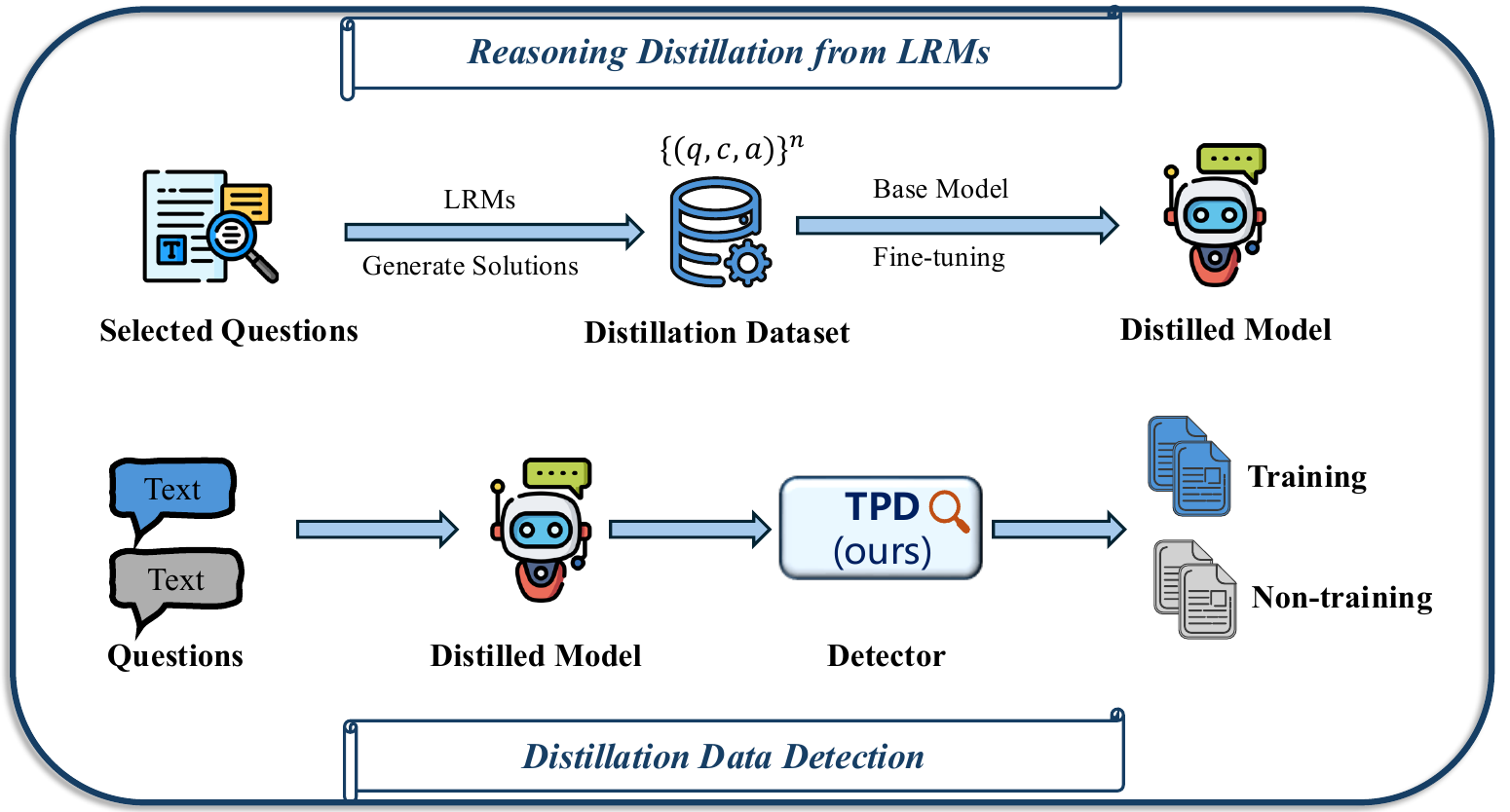}
    \caption{\textbf{Overview of distillation data detection}. The top panel illustrates the pipeline of the reasoning distillation that distills the reasoning capacities of LRMs to smaller LLMs. The bottom panel illustrates the process of detecting distillation data.}
    \label{overview}
\end{figure*}

In reasoning distillation, developers typically curate a collection of questions and employ LRMs to generate reasoning trajectories along with their answers, which subsequently serve as supervision signals for training SLMs~\citep{wu2025beyond, li2025largemodelsteachstudent}. 
However, the lack of transparency in distillation datasets raises significant concerns about benchmark contamination: evaluation data may inadvertently be included in the distillation data, leading to inflated performance metrics on corresponding benchmarks~\citep {orenproving, xu2024benchmark}. 
This underscores the importance of \textit{distillation data detection}: determining whether a given question was included in the model's distillation dataset (See Figure~\ref{overview}). 
The unique challenge of this task lies in partial availability: only the question is available at detection, without corresponding reasoning trajectories or answer. 
Accessing question-response pairs is infeasible due to post-hoc processing in solution construction~\citep{ye2025limo} and the proprietary nature of distillation datasets~\citep{guo2025deepseek, yang2025qwen3}.
This motivates us to explore an effective detection method for the question-only setting, where existing methods fail to detect distillation data given partial sample information.




In this study, we demonstrate that \textit{output tokens} generated by distilled models can expose information for identifying their distillation data. We observe that distilled models generally exhibit different probability distributions for members (\textit{i.e.}, seen questions) and non-members (\textit{i.e.}, unseen questions) in greedy decoding. In particular, distilled models tend to generate near-deterministic tokens for members, while producing more low-probability tokens for non-members. This difference in token generation behavior indicates that the probability distributions produced by distilled models can serve as membership signals for identifying distillation data.

Inspired by the analysis, we propose a simple yet effective method -- \textbf{T}oken \textbf{P}robability \textbf{D}eviation (dubbed \textbf{TPD}), which identifies the distillation data through the probabilities of output tokens generated by the model, \textit{instead of input tokens}. Our key idea behind TPD is to quantify how far the probabilities of generated tokens are from being fully deterministic. 
In particular, this can be accomplished by measuring the deviation of the generated token's probability from a high reference probability.
Consequently, our method produces smaller TPD scores for seen questions than for unseen questions during detection. 
As a result, we can achieve a clear separability of scores between seen and unseen questions, despite the partial availability setting.

Empirically, we perform extensive experiments to validate the effectiveness of the proposed method across diverse models (e.g., Qwen2.5-Instruct family~\citep{qwen2025qwen25technicalreport}, Llama-3.1-8B-Instruct~\citep{dubey2024llama}, Gemma-7B-it~\citep{team2024gemma}, and Mistral-7B-Instruct~\citep{jiang2023mistral7b}) and various datasets (e.g., S1, S1.1~\citep{muennighoff2025s} and LIMO~\citep{ye2025limo}). 
The results demonstrate that our method achieves superior performance compared to existing methods for detecting distillation data (See Table~\ref{main1}). 
For example, our method achieves an AUC of 0.918 and a TPR@1\%FPR of 0.470 on the distilled model obtained by fine-tuning Qwen2.5-32B-Instruct on the S1 dataset, indicating the effectiveness of our method for detecting distillation data. 
In addition, experimental results show that our method consistently performs well across diverse datasets and models using fixed hyperparameters, enabling deployment of our algorithm without task-specific hyperparameter tuning.
In summary, our method shows superior performance in both the AUC and TPR@1\%FPR metrics, showing the practicality of our method in real-world applications.

We summarize our main contributions as follows:
\label{Contribution}
\begin{itemize}
     \item We first present the problem of distillation data detection and emphasize its unique challenge of partial availability. We further analyze the limitations of existing methods, finding that existing methods are ineffective for detecting distillation data.
     
    \item We propose \textit{Token Probability Deviation} \textbf{(TPD)} score, a novel and effective method for detecting distillation data. The core idea of our method is to measure the deviation of \emph{\textbf{output tokens'}} probabilities produced by the distilled model from a high reference probability.

    \item Our experimental results demonstrate that \textit{output tokens} generated by distilled models can serve as effective membership signals for detecting distillation data.
    Our proposed method can effectively identify distillation data with a low false-positive rate, highlighting its practicality in real-world applications. 
\end{itemize}

\section{Preliminaries} 
\label{background}
\paragraph{Reasoning distillation}
Reasoning distillation transfers the step-by-step reasoning behavior of large reasoning models into a smaller student language model by imitating the reasoning trajectories generated by teacher models~\citep{guo2025deepseek,li2025largemodelsteachstudent}.
Let $q$ denote a question drawn from a large-scale corpus $\mathcal{Q}$, collected from diverse sources.
Using $q$ as a prompt, developers generally use LRMs to generate reasoning trajectories $c$ along with the final answer $a$. 
To construct a high-quality distillation dataset $\mathcal{D}=\{(q_i, c_i, a_i)\}_{i=1}^{N}$, developers then execute a meticulous selection process from an initial large-scale pool of candidates. 
Reasoning distillation aims to distill the reasoning capacities of LRMs to a smaller model by self-supervised fine-tuning an SLM on the resulting distillation dataset (See Figure~\ref{overview}).
Formally, the objective of training can be formulated as:
\begin{equation}
\mathcal{L}_{\theta} = -\sum_{t=1}^{N} \log P_{\theta} \big( y_t \mid y_{<t}, q \big),
\end{equation}
where $q$ denotes the input question, and $y = \{y_1, y_2, \ldots, y_N\}$ represents the corresponding target sequence, comprising the reasoning trajectory $c$ and the final answer $a$. 
$P_{\theta} ( y_t \mid y_{<t}, q )$ represents the conditional probability of generating token $y_t$, given the query $q$ and the preceding sequence $y_{<t}$.
This paradigm explicitly trains the student model on question-response pairs generated by large reasoning models (See Appendix~\ref{example_train} for an example), thereby enabling the model to internalize intermediate reasoning patterns of teacher models.

\paragraph{Membership inference} Membership Inference Attacks (MIAs) aim to determine whether a particular record is included in the training data of a model~\citep{shokri2017membership}. 
Given a trained model $f(\boldsymbol{x}, \boldsymbol{\theta})$ and a data point $(\boldsymbol{x}, y)$, traditional MIAs infer whether a target data point belongs to the training dataset $\mathcal{D}_{\mathrm{train}}$. Training data detection for LLMs is an instance of membership inference attacks~\citep{shidetecting}. The existing MIAs on large language models usually aim to infer whether a given piece of text $\boldsymbol{x}$ belongs to the training corpus of a large language model $\mathcal{M}$.
Existing detection methods for LLMs typically use a scoring function based on the likelihood of \textit{\textbf{input sequence}} to detect training data~\citep{li2023estimating, shidetecting, zhang2024min}. 
For example, \textit{Perplexity} is proposed to distinguish members and non-members, based on the observation that members tend to have lower perplexity than non-members~\citep{li2023estimating}. Formally, The perplexity of $\boldsymbol{x}$ is calculated as:
\begin{equation}
\text{Perplexity}(\boldsymbol{x}; \mathcal{M}) = \exp \{ - \frac{1}{n} \sum_{i=1}^{n} \log p_{\boldsymbol{\theta}}(x_i \mid x_1, \dots, x_{i-1}) \},
\end{equation}
where $\boldsymbol{x} = \{x_1, x_2, \dots, x_n$\} is a sequence of input tokens and $p_{\boldsymbol{\theta}}(x_i \mid x_1, \dots, x_{i-1})$ is the conditional probability of $x_i$ given the preceding tokens. Although some methods for pretraining and fine-tuning data detection have been studied, membership inference on distillation data for reasoning distillation remains underexplored. In the next section, we introduce the distillation data detection task, a tailored formulation of this problem.

\section{Distillation Data Detection}
\label{setting}
In this section, we formally define the \emph{\textbf{Distillation Data Detection}} task, which is uniquely challenging due to the partial availability of distillation data. 
The goal of our task is to predict whether a given question is included in the model's distillation dataset. 

\paragraph{Problem definition}
Using question $q$ as a prompt, developers often generate distillation data by sampling responses from multiple advanced LRMs and by refining them to obtain high-quality reasoning trajectories $c$ and corresponding answer $a$~\citep{ye2025limo, wu2025beyond, tian2025beyond, zhuang2025unicott}. 
However, the resulting distillation dataset $\mathcal{D}=\{(q_i,c_i,a_i)\}_{i=1}^{N}$ is often proprietary~\citep{guo2025deepseek, yang2025qwen3}---\textit{i.e.}, the exact reasoning trajectory and answer are inaccessible for a given question. 
Additionally, due to the non-deterministic generation process and post-hoc filtering, it is generally infeasible to recover the exact reasoning trajectory or answer associated with a given question. 
Thus, we study a more practical \emph{question-only} setting in which an auditor can query a distilled model $\mathcal{M}$ with question $q$ and obtain model outputs, but has no access to the corresponding reasoning trajectories $c$ and answer $a$ of a datapoint.

Formally, let $\mathcal{Q}_{\mathcal{D}}=\{q_i:(q_i,c_i,a_i)\in\mathcal{D}\}$ denote the set of questions from a distillation dataset used for training student models.
We pose distillation data detection as a level-set estimation problem defined on a scoring function $\mathcal{S}(q,  \mathcal{M})$ as:
\begin{equation}
\label{eq:estimator}
G(q; \mathcal{M}) =
\begin{cases} 
\text{1} & \text{if } \mathcal{S}(q,  \mathcal{M}) < \lambda, \\
\text{0} & \text{if } \mathcal{S}(q,  \mathcal{M}) \geq \lambda,
\end{cases}
\end{equation}
where $G = 1$ indicates \textit{member}~($q\in\mathcal{Q}_{\mathcal{D}}$) and $G = 0$ indicates \textit{non-member}~($q\notin\mathcal{Q}_{\mathcal{D}}$), with $\lambda$ being a case-dependent threshold.
The key difficulty of this task lies in \emph{\textbf{partial availability}}: the training datapoint is triple $x=(q,c, a)$, yet only the question $q$ is available at test time. Consequently, the design of $\mathcal{S}$ must rely solely on question-conditioned behaviours of $\mathcal{M}$, rather than on likelihood-based metrics over the ground truth $(c, a)$.

\paragraph{Challenge of partial availability}
Most prior work on training data detection for LLMs assumes access to the \emph{entire} training sample, which contains complete information seen during training~\citep{mattern2023membership, fu2024membership, mireshghallah2022quantifying}.
In this setting, the scoring function $\mathcal{S}$ can be defined directly in terms of sample likelihoods, exploiting probability estimates over input tokens.
Existing training data detection approaches targeting LLMs typically leverage the likelihood of \textit{input tokens} to design a scoring function for detection.
For example, \textsc{Min-K\%}~\citep{shidetecting} computes the average log-likelihood of the lowest k\% tokens scores over the input sequence, effectively using likelihood  of input tokens as membership signals. 
Such approaches are well-suited when full sample information is observable during detection.

\begin{wrapfigure}{r}{0.4\textwidth}
\vspace{-15pt}
    \centering
    \includegraphics[width=0.35\textwidth]{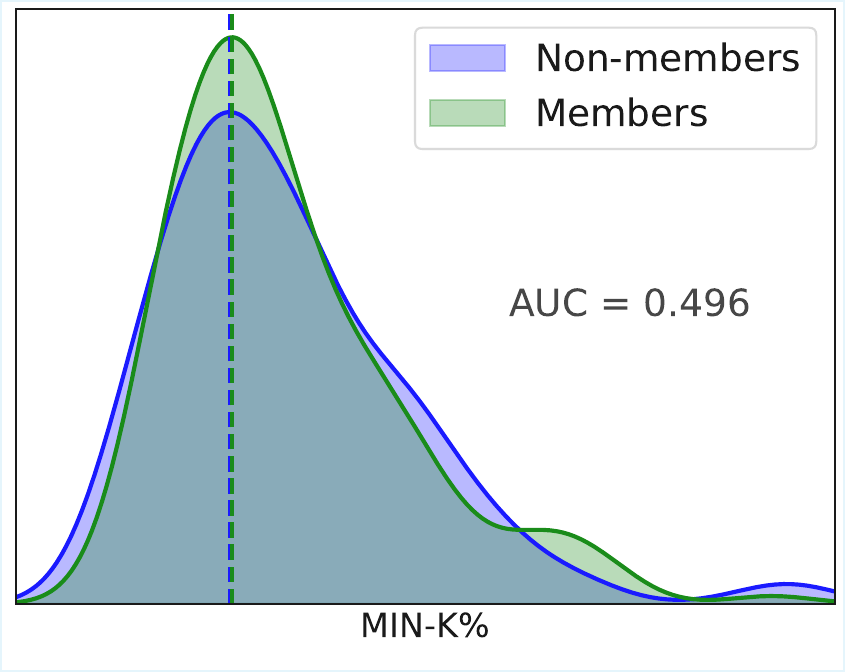}
    \vspace{-10pt}
    \caption{Scores distribution of Min-K\% for members and non-members from the distilled model trained on the LIMO dataset using Qwen2.5-32B-Instruct.}
    \label{score_dis}
\vspace{-10pt}
\end{wrapfigure}

However, these approaches may perform poorly in distillation data detection when only partial sample information is available.
The absence of joint question-response pairs weakens the key signal exploited by likelihood-based approaches operating over input sequences, leaving them ill-suited for this task. 
To illustrate, we analyze the distribution of \textsc{Min-K\%} scores for member versus non-member questions. 
Figure~\ref{score_dis} shows that the two distributions exhibit substantial overlap, indicating limited separability and poor discriminative power when only questions are available. 
Furthermore, in Appendix~\ref{evaluation_partial}, we evaluate existing methods under two detection scenarios: using only the question and using the question–response pair. 
The results indicate that existing methods are ineffective in the question-only setting but perform well when the full question–response pair is available, demonstrating that partial availability undermines the effectiveness of current approaches.
This highlights the need for alternative effective scoring functions under partial availability. 
In light of the challenge of partial availability, we investigate whether the \emph{token generation behavior} of distilled models, conditioned solely on $q$, can serve as a reliable signal of membership.

\section{Method}
\label{sec:method}
To address the challenge of partial availability, we explore a question-only scoring approach that leverages the token-level generation behavior of distilled reasoning models for detection.
We begin by comparing the probability patterns of generated tokens between member and non-member questions~(See Section~\ref{motivating_analysis}).
Subsequently, building on our empirical observations, we propose \emph{Token Probability Deviation}~(\textbf{TPD}), a novel and effective method to detect distillation data from large language models~(See Section~\ref{tbd}).

\begin{figure*}[t!]
    \label{motivation}
    \centering
    \begin{subfigure}[b]{0.6\textwidth}
        \centering
      \includegraphics[width=1.0\textwidth, height=0.25\textheight, keepaspectratio=False]{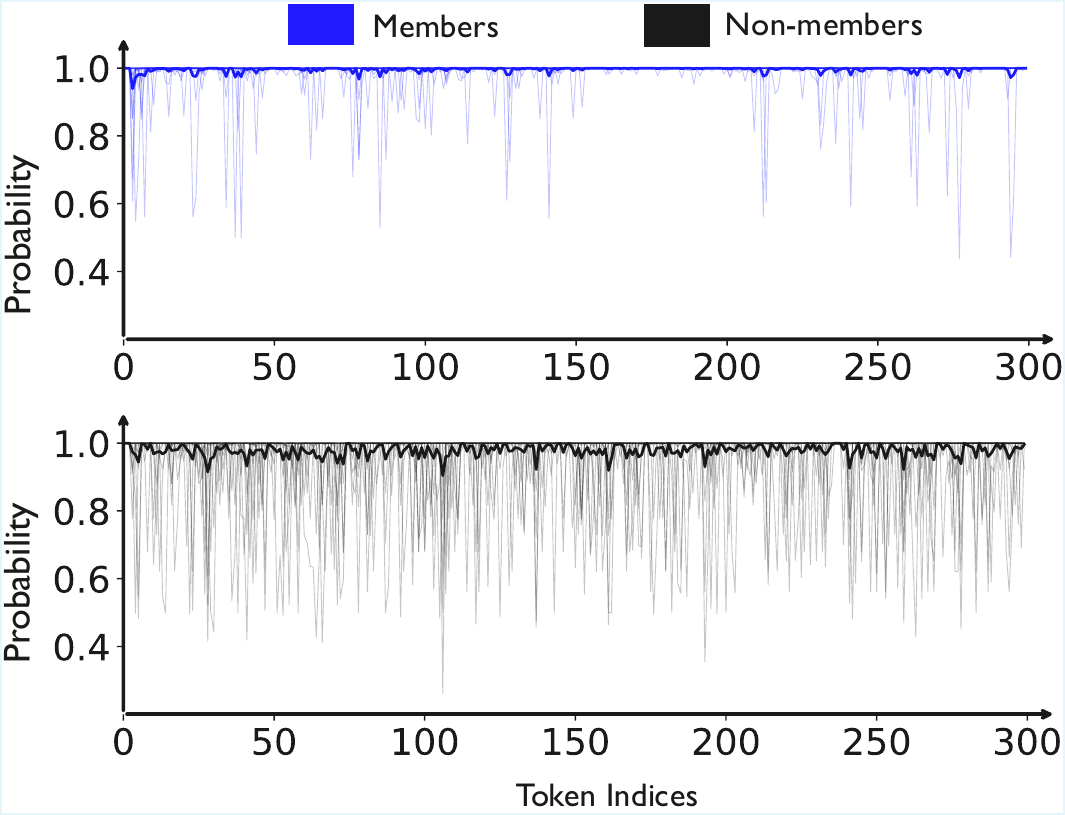}
        \caption{Token-wise probability distributions}
        \label{motivation_a}
    \end{subfigure}
    \hspace{0.0cm}
    \begin{subfigure}[b]{0.38\textwidth}
        \centering
        \includegraphics[width=1.0\textwidth, height=0.25\textheight, keepaspectratio=False]{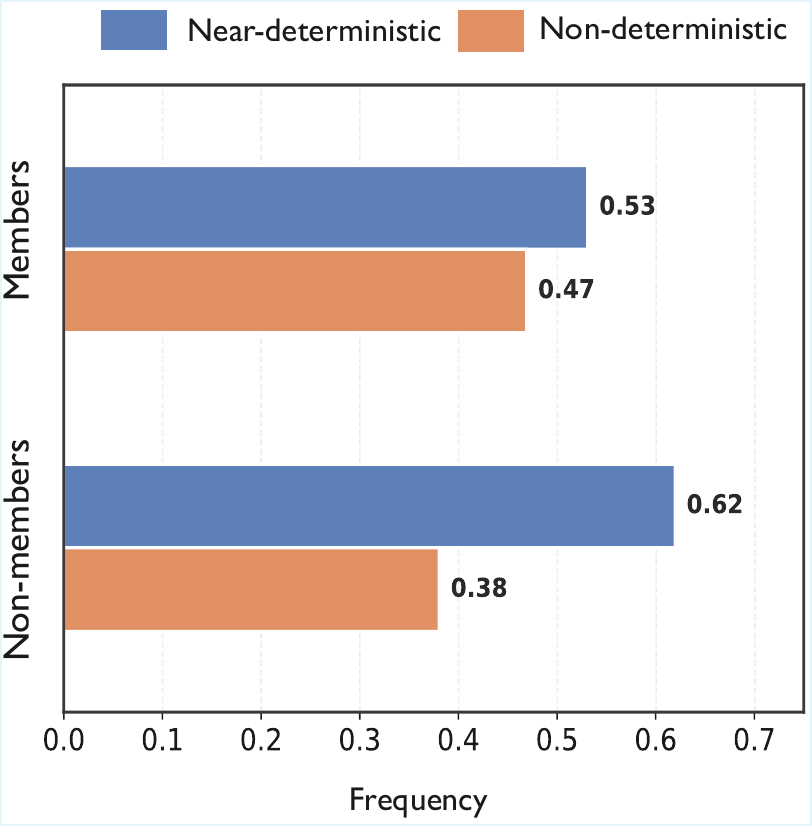}
        \caption{Near-deterministic tokens comparison}
        \label{motivation_b}
    \end{subfigure}
    \caption{Token-level generation behavior of distilled models on member vs. non-member questions under greedy decoding. \textbf{(a) Token-wise probability distributions}: we contrast the distribution of token-wise probability between members and non-members, showing that non-members incur more generated tokens with lower probability. \textbf{(b) Near-deterministic vs. non-deterministic tokens}: near-deterministic tokens denote generated tokens with probabilities approaching 1, and vice versa for non-deterministic tokens. We observe that the distilled model tends to generate more near-deterministic tokens for members.}
\end{figure*}

\subsection{Motivating Analysis} 
\label{motivating_analysis}
\paragraph{Analysis setup}
The goal of our analysis is to investigate whether the probability patterns of output tokens produced by a distilled model differ between member and non-member questions. 
Following prior work~\citep{muennighoffs1}, we first distill the reasoning capabilities from LRMs to the Qwen2.5-32B-Instruct model via supervised full-parameter fine-tuning on the S1 dataset.
The dataset is split into training and testing subsets with an 8:2 ratio, from which we sample members~(training set) and non-members~(testing set).
This ensures an i.i.d setup, with both groups drawn from the same underlying data distribution. 
For each question, we generate a response from the distilled model using greedy decoding and extract the token probabilities of the response for comparison. In Appendix~\ref{sample_ex}, we provide illustrative prompts.

\paragraph{Members induce near-deterministic \textit{output tokens} more frequently}
To examine distributional differences in output token probabilities, we analyze sequences of up to 300 output tokens generated by the model for both member and non-member questions. 
Figure~\ref{motivation_a} shows the token-wise maximum probability distributions from the distilled model across 20 member and 20 non-member samples. 
The horizontal axis denotes the token index, and the vertical axis reports the probability assigned to each token. 
Thin lines correspond to individual samples, and bold lines denote mean token-wise probability across all members or non-members.
We observe that the distilled model tends to frequently generate tokens with probabilities close to~1 for members,  while producing more low-probability tokens for non-members.
Figure~\ref{motivation_b} further contrasts the frequency of near-deterministic tokens~(with probability approaching 1) against non-deterministic tokens. 
The results show that the distilled model produces a substantially higher fraction of near-deterministic tokens for member questions. 
This suggests that the probabilities of output tokens likely carry membership signals, motivating our design of a scoring function that leverages \textit{output tokens} generated by the distilled model to detect distillation data.

\subsection{Token Probability Deviation Score} 
\label{tbd}
Motivated by our preliminary analysis, we propose Token Probability Deviation~(TPD) score, a method that exploits the observation that member questions tend to elicit near-deterministic tokens, whereas non-members induce relatively more low-probability tokens. Our TPD score utilizes the \emph{output} tokens' probabilities along the model’s reasoning trajectory for a given question $q$. This design sidesteps the partial-availability constraint of distillation datasets while providing a simple, model-agnostic signal for distillation data detection.

The core idea of our method is to measure the deviation of the generated tokens' probability from a high reference probability.
Given a question with a sequence of tokens $q = \{q_1, q_2, ..., q_N\}$, the tokens generated by the model can be denoted as $y=\{y_1, y_2, ..., y_i\}$. 
We use $p_{\theta}(y_i \mid y_{<i}, q)$ to denote the probability that the target model predicts $y_i$, given the question prompt $q$ and the generated text prefix $y_{<i} = \{y_1, y_2, ..., y_{i-1}\}$. 

Since distilled models tend to generate tokens with extremely high probability for seen questions, outlier tokens are likely to display a highly distinctive membership signal.
This motivates us to quantify the deviation between the probability of the generated token $y_i$ and the high-confidence reference probability $\tau$. Formally, we define the deviation term as:
\begin{equation}
    d_i(q; \tau) = \max\bigl(0, \tau - p_{\theta}(y_i \mid y_{<i}, q)\bigr),
\end{equation}
where $\max(0, \cdot)$ ensures that only generated tokens whose probabilities fall below the reference probability threshold $\tau$, contribute to the sample score. 

Empirically, we observe that earlier generated tokens carry more distinct membership signals than subsequent ones, as shown in Appendix~\ref{length_compare}. Consequently, we focus on the first $M$ generated tokens to obtain distinguishable membership signals. 
The final robust sample score is obtained by averaging the token-level deviations $ d_i(q; \tau)$. Concretely, our TPD score can be formulated as:
\begin{equation}
\label{method_ep}
    \mathcal{S}(q,\theta) = \frac{1}{E} \sum_{i=1}^{M}d_i(q;\tau)^{\alpha},
\end{equation}
where $E = \sum_{i=1}^{M} \mathbf{1}\!(p_{\theta}(y_i \mid y_{<i}, q)<\tau)$ denotes the number of outliers among the first $M$ tokens, and $\alpha$ denotes a parameter to control token-level deviations that contribute to the final score.
For instance, a small value of $\alpha$ (e.g., 0.6) emphasizes the role of smaller token-level deviation in the final score.
Our experimental results in Figure~\ref{alpha} show that applying $\alpha$ can yield an improved TPR@1\% FPR for detecting distillation data.

\paragraph{Detection with our TPD}
Our method enables us to build a detector $G(q; M)$ for a distilled model to infer the membership of a given question $q$. 
Notably, our method achieves strong detection performance across diverse datasets under a unified hyperparameter configuration, requiring no task-specific tuning.
At test time, samples with lower scores $\mathcal{S}(q,\theta)$ are classified as distillation data and vice versa. 
A case study in Appendix~\ref{case_study} further validates the practical utility of our TPD, achieving a high detection accuracy of 0.86.
Additionally, as shown in Appendix~\ref{overhead}, our method incurs minimal computational overhead, making it practical for real-world deployment.

\section{Experiments}
In this section, we evaluate the performance of our method across several datasets using multiple models of varying sizes. Extensive experiments demonstrate the effectiveness of our method, showing its capability to detect distillation data in a practical scenario.

\subsection{Experimental Setup}
\label{exp:setup}
\paragraph{Datasets and models}
We conduct experiments on several high-quality distillation datasets provided by previous work, including S1, S1.1~\citep{muennighoffs1}, and LIMO~\citep{ye2025limo}. The details of the datasets are provided in Appendix~\ref{details of dataset}. Specifically, we fine-tune the base model (e.g, Qwen2.5-32B-Instruct) on these datasets with a full-parameter supervised fine-tuning strategy. In addition, we also perform experiments on various models, such as Qwen2.5-Instruct family~\citep{qwen2025qwen25technicalreport}, Llama-3.1-8B-Instruct~\citep{dubey2024llama}, Gemma-7B-it~\citep{team2024gemma}, and Mistral-7B-Instruct~\citep{jiang2023mistral7b} models.

\paragraph{Baseline methods} We compare our method against current baselines for LLMs. Specifically, \textit{Input-token-based methods} compute membership scores based on the likelihood of input tokens, including \textbf{Perplexity}~\citep{li2023estimating}, 
\textbf{Zlib}, \textbf{Lowercase}~\citep{carlini2021extracting}, 
\textbf{Neighbor}~\citep{mattern2023membership}, 
\textbf{Min-K\%}~\citep{shidetecting}, 
\textbf{Min-K\%++}~\citep{zhang2024min}, 
and \textbf{Infilling Score}~\citep{raoof2025infilling}. 
\textit{Output-token-based methods} leverage tokens generated by the model as the detection signal, including \textbf{Entropy-Temp}, \textbf{Entropy-Noise}, 
and \textbf{Self-Critique}~\citep{tao2026detecting}.  We provide the details of each baseline in Appendix~\ref{baseline}.
Additionally, we introduce two naive variants of our method as ablation baselines: \textbf{Generated Perplexity} that computes the perplexity using the probabilities
of generated tokens, and \textbf{Generated Min-K\%} that computes the average log-likelihood of the $K\%$ generated tokens with the lowest predicted probability. 

\paragraph{Implementation details}
In our experiments, we fine-tune various models (e.g., Qwen2.5-32B-Instruct) separately on several distillation datasets to obtain different distilled models. 
To ensure a fair evaluation, all experiments use our own distilled models.
Training and test sets are drawn from the same dataset to avoid distribution shift, which may lead to unreliable evaluation results. 
For main experiments, we generate responses from distilled models under greedy decoding, and compute TPD with hyperparameters $M=300$, $\tau=1$, and $\alpha=0.6$. More details of the implementation and evaluation metrics are provided in Appendix~\ref{implementation}.
\begin{table*}[!t]
\centering
\footnotesize
\caption{AUC of our method and baselines on diverse distilled models. These models are produced through fine-tuning different-sized models (e.g., Qwen2.5-32B-Instruct) on various distillation datasets, including S1, LIMO, and S1.1 datasets. \textbf{Bold} shows the superior result.}
\label{main1}
\renewcommand\arraystretch{1.1}
\resizebox{1.0\textwidth}{!}{
\setlength{\tabcolsep}{1.5mm}{
\begin{tabular}{lccc|ccc|ccc}
\toprule
\multirow{2}{*}{\textbf{Method}}& \multicolumn{3}{c}{Qwen2.5-7B-Instruct} & \multicolumn{3}{c}{Qwen2.5-14B-Instruct} &\multicolumn{3}{c}{Qwen2.5-32B-Instruct}\\
\cmidrule(lr){2-4} \cmidrule(lr){5-7} \cmidrule(lr){8-10} 
 & \multicolumn{1}{c}{S1} & \multicolumn{1}{c}{LIMO} & \multicolumn{1}{c}{S1.1} & \multicolumn{1}{c}{S1} & \multicolumn{1}{c}{LIMO} & \multicolumn{1}{c}{S1.1} & \multicolumn{1}{c}{S1} & \multicolumn{1}{c}{LIMO} & \multicolumn{1}{c}{S1.1}\\
\midrule
\textit{Input-token-based methods} \\
Perplexity~\citep{li2023estimating}  &0.444	&0.482	&0.503 &0.449	&0.498	&0.517 &0.433	&0.499	&0.487\\
Lowercase~\citep{carlini2021extracting} &0.435	&0.472	&0.493 &0.467	&0.507	&0.489 &0.459	&0.475	&0.463\\
Zlib~\citep{carlini2021extracting} &0.474	&0.486	&0.467 &0.467	&0.495	&0.471 &0.448	&0.496	&0.447\\
Neighbor~\citep{mattern2023membership} &0.539	&0.503	&0.441	&0.543	&0.500	&0.435	&0.555	&0.503	&0.444\\
\textsc{MIN-K\%}~\citep{shidetecting} &0.443	&0.480	&0.494 &0.453	&0.496	&0.509 &0.437	&0.496	&0.479\\
\textsc{MIN-K\%++}~\citep{zhang2024min} &0.472	&0.458	&0.486 &0.509	&0.508	&0.489 &0.461	&0.461	&0.439\\
Infilling Score~\citep{raoof2025infilling} &0.529	&0.529	&0.520	&0.534	&0.544	&0.493	&0.574	&0.489	&0.475 \\

\midrule
\textit{Output-token-based methods} \\
Entropy-Temp~\citep{tao2026detecting} &0.552	&0.476	&0.450	&0.581	&0.475	&0.415	&0.703	&0.493	&0.437\\
Entropy-Noise~\citep{tao2026detecting} &0.511	&0.495	&0.535	&0.470	&0.498	&0.561	&0.460	&0.494	&0.566\\
Self-Critique~\citep{tao2026detecting} &0.436	&0.502	&0.475	&0.404	&0.477	&0.475	&0.440	&0.504	&0.466\\
Generated Perplexity &0.753	&0.605	&0.564 &0.785	&0.596	&0.558 &0.847	&0.662	&0.619\\
Generated \textsc{Min-k} &0.754	&0.604	&0.563 &0.785	&0.596	&0.559 &0.847	&0.661	&0.619\\
\rowcolor{gray!10}
Ours &\textbf{0.855}	&\textbf{0.694}	&\textbf{0.617}  &\textbf{0.870} &\textbf{0.671} &\textbf{0.562} &\textbf{0.918} &\textbf{0.728} &\textbf{0.649}\\
\bottomrule
\end{tabular}
}}
\end{table*}

\paragraph{Is our method effective across models trained on various datasets?}
To investigate the performance of baselines and our method across diverse distilled models, we fine-tune diverse models on various distillation datasets, including S1, LIMO, and S1.1 datasets. 
Table~\ref{main1} shows that our method significantly outperforms the baselines, achieving superior performance for detecting distillation data. 
Table~\ref{main1_tpr} in Appendix~\ref{result} shows that our method achieves a higher TPR@1\%FPR score compared to all baselines.
For example, our method achieves a high AUC of 0.918 and a TPR@1\% FPR of 0.470 on the distilled model obtained by fine-tuning Qwen2.5-32B-Instruct on S1.
Our experiments demonstrate that tokens generated by distilled models can serve as effective membership signals for detecting distillation data.
Additionally, the results in Appendix~\ref{data_size} demonstrate that our method consistently achieves reliable detection performance across various sizes of distillation datasets.
We further evaluate our method on the large-scale dataset OpenR1-Math-220K, with results reported in Appendix~\ref{exp_larger}, where our method achieves superior performance.
Overall, these findings highlight the effectiveness of our approach to distillation data detection across various models and datasets.

\paragraph{Is our method effective across models from different families?}
To validate the effectiveness of our method across models with different architectures, we conduct experiments on the S1 dataset with three different language models, including Llama-3.1-8B-Instruct~\citep{dubey2024llama}, Gemma-7B-it~\citep{team2024gemma}, and Mistral-7B-Instruct-v0.3~\citep{jiang2023mistral7b} models. Table~\ref{various_model} in Appendix~\ref{result} reports the AUC score of our method and baselines across various models. The results show that our method significantly outperforms the baselines, which are ineffective for detecting distillation data. Overall, our experiments demonstrate that our method is model-agnostic and broadly applicable for detecting distillation data.

\begin{figure*}[t!]
    \centering
    \begin{subfigure}[t]{0.31\textwidth}
        \centering
      \includegraphics[width=\linewidth]{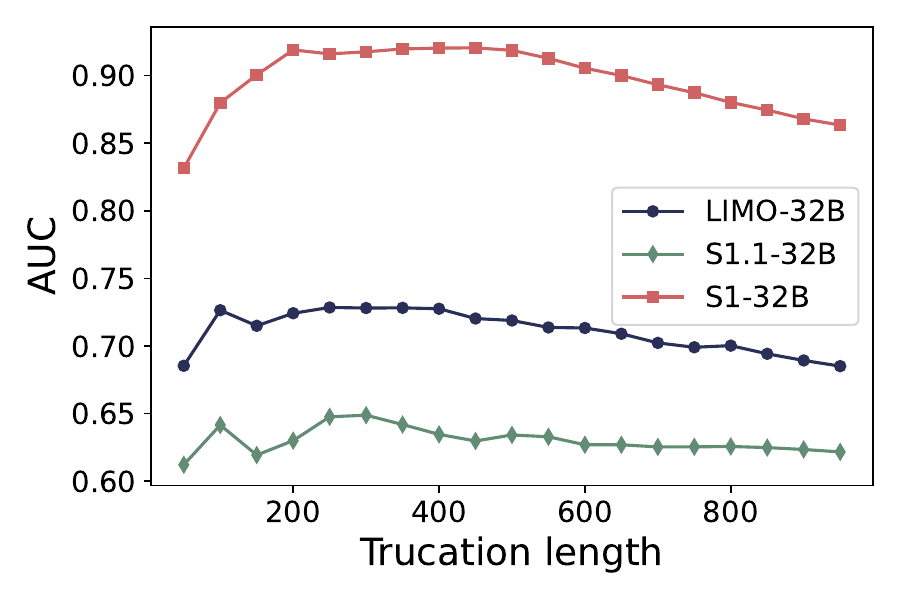}
        \caption{AUC score vs. $M$}
        \label{auc_length}
    \end{subfigure}
    \hspace{0.0cm} 
    \begin{subfigure}[t]{0.33\textwidth}
        \centering
        \includegraphics[width=\linewidth]{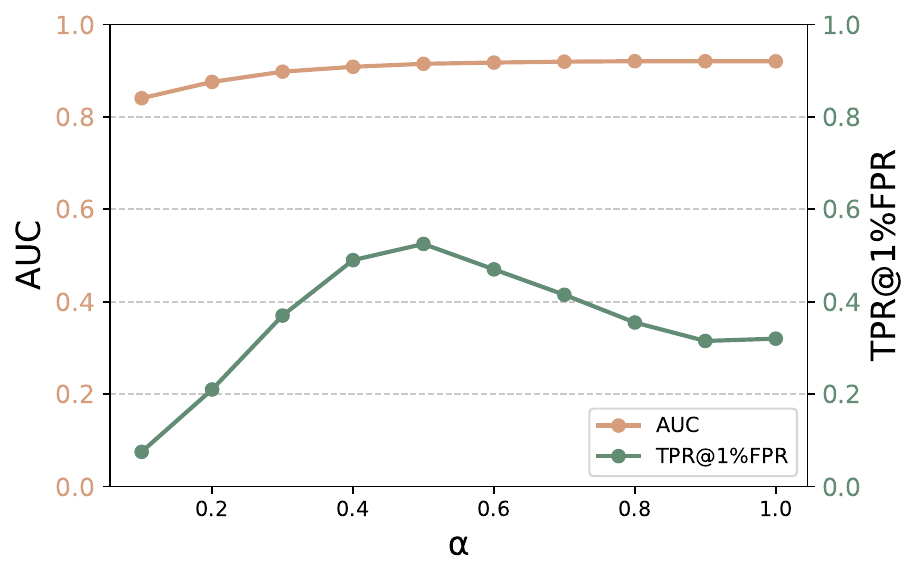}
        \caption{AUC and TPR scores vs. $\alpha$}
        \label{alpha}
    \end{subfigure}
    \hspace{0.2cm} 
        \begin{subfigure}[t]{0.30\textwidth}
        \centering
        \includegraphics[width=\linewidth]{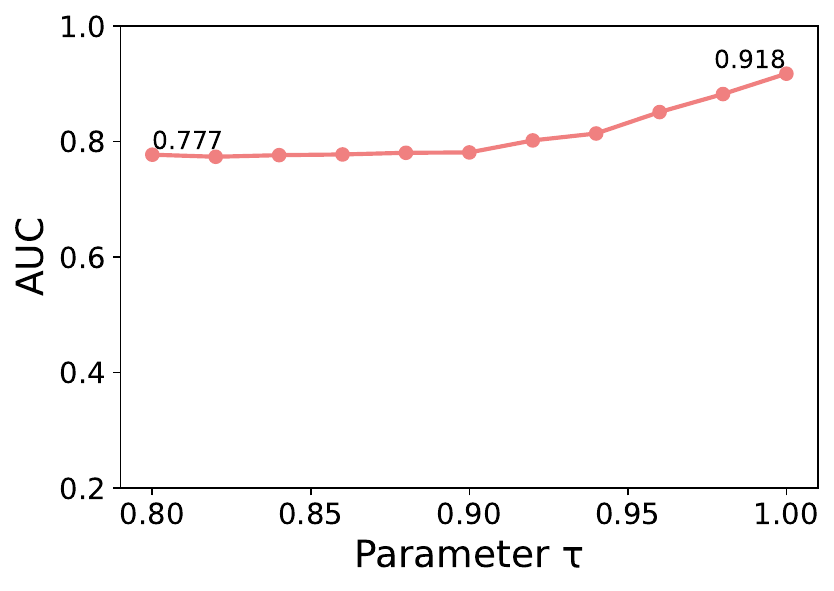}
        \caption{AUC score vs. $\tau$}
        \label{tau}
    \end{subfigure}
    \caption{Effect of truncation length $M$ (\ref{auc_length}) and parameter $\alpha$ (\ref{alpha}) and reference probability $\tau$ (\ref{tau}) on the performance of our method.}
    \label{tokens_numbers}
\end{figure*}

\paragraph{Hyperparameter analysis.}
We perform experiments with varying $M$, $\alpha$, and $\tau$ values to examine their effect on our method's performance, using the distilled model obtained by fine-tuning Qwen2.5-32B-Instruct on S1. 
(1) Effect of $M$: Figure~\ref{auc_length} shows that the AUC initially increases with truncation length, but declines beyond a certain point, suggesting that excessive outputs introduce noise that weakens the membership signal. 
Our method consistently achieves near-optimal performance across all datasets at a truncation length of approximately 300, demonstrating its robustness and allowing users to deploy our algorithm with a fixed truncation length. 
(2) Effect of $\alpha$: Figure~\ref{alpha} shows the AUC and TPR@1\%FPR scores of our method with varying $\alpha$. 
Increasing $\alpha$ initially improves the AUC score, and performance stabilizes as $\alpha$ continues to increase. 
The TPR@1\%FPR score increases significantly as $\alpha$ increases, reaching a peak near $\alpha=0.6$ and subsequently decreases. 
This behavior allows us to deploy our algorithm flexibly by simply adjusting $\alpha$, targeting the preferred metric in practical applications. 
(3) Effect of $\tau$: Figure~\ref{tau} shows our method achieves superior performance with a large value of $\tau$.  As described in Section~\ref{motivating_analysis}, the distilled reasoning model tends to generate tokens with extremely high probability. Thus, applying a high reference probability can help identify outlier tokens and improve performance in detecting distillation data.

\begin{wraptable}{r}{0.6\textwidth}
\centering
\vspace{-10pt}
\footnotesize
\caption{Ablation study on the components of our method.}
\label{ablation}
\renewcommand\arraystretch{1.1}
\resizebox{0.6\textwidth}{!}{
\setlength{\tabcolsep}{1mm}{
\begin{tabular}{ccc|cc}
\toprule
Truncation length  &Token deviation  & Parameter $\alpha$ & AUC & TPR@1\% FPR \\
\midrule
\ding{55} & \ding{55}  &  \ding{55}       & 0.847 & 0.170\\
$M=300$ &  \ding{55}     & \ding{55}        & 0.903 & 0.290 \\
$M=300$ &  $\checkmark$    &\ding{55} & 0.920 & 0.320\\
\rowcolor{gray!10}
$M=300$ &  $\checkmark$ & $\alpha=0.6$ &  \cellcolor{gray!10}\underline{0.918} & \cellcolor{gray!10} \textbf{0.470} \\
\rowcolor{gray!10}
$M=300$ & $\checkmark$ & $\alpha=0.7$ & \textbf{0.920} & \underline{0.415} \\
\bottomrule
\end{tabular}
}}
\vspace{-10pt}
\end{wraptable}

\paragraph{Decomposing the contribution of our method.}
Our method can be decomposed into three key components: \textbf{(1)} truncation operation, which truncates the generated sequence to the first $M$ tokens; \textbf{(2)} token deviation measure $d_i(q;\tau)$, which measures the deviation of the generated token's probability from a reference probability $\tau$; and \textbf{(3)} a scaling parameter $\alpha$, which controls how token-level deviations contribute to the sample-level score.
To elucidate individual contributions of each component, we conduct an ablation study in Table~\ref{ablation}.
We begin with a simple baseline that averages predicted probabilities over the first 1000 generated tokens.
We then gradually incorporate the above components into the score computation to obtain the final formulation of our method. 
Firstly, truncating the generated output to 300 tokens for scoring yields a substantial performance gain.
Secondly, incorporating the token deviation measure further improves performance, confirming that focusing on outlier tokens produces a more discriminative membership signal. 
Finally, introducing a scaling parameter $\alpha$ to modulate token-level contributions leads to a significant improvement in TPR@1\%FPR. 
By combining these components, our method achieves superior performance in both AUC and TPR@1\% FPR scores.

\paragraph{How does training configuration affect our method's performance?}
To investigate how training configurations affect our method's performance, we conduct experiments on the S1 dataset using Qwen2.5-7B-Instruct across varying training parameters, including learning rate (e.g. $1 \times 10^{-6}$, $1 \times 10^{-5}$, $1 \times 10^{-4}$) and epoch (e.g. 3, 4, 5).
Table~\ref{tab:training_setups} in Appendix~\ref{training_configuration} reports the AUC scores of baselines and our method across different training configurations.
The results show that our method consistently outperforms all baselines, demonstrating its effectiveness across various training setups.
We observe that the performance of our method improves with more epochs, whereas a learning rate of $1\times 10^{-6}$ is too small for the model to converge, leading to suboptimal results.
Notably, our training configuration (learning rate of $1 \times 10^{-5}$ and 5 epochs) aligns with common distillation practice~\citep{muennighoffs1}, where our method achieves strong performance.
Overall, the results show that our method is applicable to real-world reasoning distillation scenarios.

{\section{Discussion}}
\paragraph{Are machine-generated text detection methods applicable to detecting distillation data?}
Recent work examines the transferability between membership inference attacks and machine-generated text detection, indicating that methods designed for machine text detection can be used to identify the membership of data~\citep{koike2025machine}.
To investigate the effectiveness of machine-generated text detection methods in distillation data detection, we compare our method with four methods of machine-generated text detection: DetectGPT~\citep{mitchell2023detectgpt}, Fast-DetectGPT~\citep{bao2024fastdetectgpt}, Binoculars~\citep{hans2024spotting}, and DetectLLM~\citep{su2023detectllm}. 

Table~\ref{ai_gen} in Appendix~\ref{result} shows that existing methods on machine text detection fail to detect distillation data, while our method achieves superior performance. Notably, our approach exploits the model's token-level generation probabilities for a given input question, distinguishing it from existing machine-generated text detection and MIAs methods that both rely exclusively on signals extracted from the input text itself. Consequently, machine-generated text detection approaches are unable to overcome the challenge of partial availability in distillation data detection. 
The experimental results demonstrate that existing methods for machine-generated text detection fail to generalize to distillation data detection.

\paragraph{Is our method affected by decoding sampling parameters?}
Our method is motivated by the observation that, under greedy decoding, distilled models generally exhibit different probability distributions for members and non-members. 
A natural question is: \textit{how does our method perform under different sampling parameters?}
To investigate the performance of our method with different decoding parameters, we conduct experiments on the S1 dataset using Qwen2.5-7B-Instruct with varying temperatures and Top-K values.
In the experiment on the effect of temperature, top-k was set to 30 to isolate the effect of the varying parameter. Similarly, in the top-k experiment, the temperature was kept constant at 0.3. 
Table~\ref{tab:decoding} in Appendix~\ref{parameters} presents the AUC score of our method with different sampling parameters. The experimental results show that our method can achieve better performance with a lower temperature, while its performance remains stable across varying Top‑K values.
Notably, our method achieves superior performance under greedy decoding, which aligns with the design rationale of our approach.

\paragraph{How effective is our method against question paraphrasing?} 
To examine the effectiveness of our method in a scenario where reasoning distillation is applied with paraphrased questions to evade detection, we conduct experiments on three different models using the S1 dataset. Concretely, we use GPT-5-mini\footnote{\url{https://platform.openai.com/docs/models/gpt-5-mini}} to paraphrase the original question, obtaining a rephrased version that remains semantically consistent with the original question. 
We then evaluate our method on paraphrased questions, simulating an evasion scenario in which the questions used for reasoning distillation are unavailable due to intentional paraphrasing.
Table~\ref{paraphrase} in Appendix~\ref{result} reports the AUC scores of our method and baselines.
The results show that our method consistently outperforms baselines, indicating its capability to detect distillation data in the question paraphrasing scenario.

\section{Conclusion}
In this work, we first formalize the problem of distillation data detection and emphasize its unique challenge of partial availability. Our empirical analysis shows that existing methods are ineffective for distillation data detection.
We then propose \textit{Token Probability Deviation} (\textbf{TPD}) score, a novel and effective method for detecting distillation data, which quantifies the deviation of \textit{output tokens'} probabilities from a high reference probability.
In this work, we demonstrate that output tokens produced by distilled models can serve as effective information for identifying membership of data in distillation data detection.
Experimental results show that our method significantly outperforms current baselines across diverse datasets and model families for detecting distillation data.
Our TPD can identify distillation data with a low false-positive rate, highlighting its practicality in real-world applications. 
In addition, the experiments show that our method is robust to parameter choice, enabling us to deploy our algorithm without task-specific hyperparameter tuning. 
In summary, we introduce a detection method that achieves superior performance in identifying distillation data used for reasoning distillation.
We hope that our study can advance further research on data contamination resulting from reasoning distillation.

\paragraph{Limitations}
Our work focuses on detecting training data used in reasoning distillation. 
The proposed method is not designed for detecting pretraining data, and consequently exhibits weaker performance in that setting.
Thus, developing a unified approach that generalizes across all training data detection tasks remains an important direction for future work.

\clearpage

\bibliographystyle{plain}

\bibliography{refs}



\appendix
\clearpage

\addcontentsline{toc}{section}{Appendix}
\part{Appendix} 
\parttoc 



\section{Related Work}
In this paper, we propose a problem about distillation data detection, which is related to an amount of literature on reasoning distillation and detecting training data from Large Language Models. We discuss related works in two directions relevant to our study.

\paragraph{Distilling reasoning capability from LRMs.}
Large Reasoning Models (LRMs) exhibit remarkable performance in solving complex tasks, achieving this by training the model to produce a long chain of thought reasoning process before responding with the final answer~\citep{jaech2024openai, yang2025qwen3, comanici2025gemini}. However, developing reasoning models, which achieve reasoning capacity compared to large reasoning models, remains a significant challenge for the research community~\citep {kumartraining, xin2024deepseek, wan2024alphazero}. Recently, a growing literature has focused on reasoning distillation, which improves the reasoning capabilities of models with lower computational cost~\citep{guo2025deepseek, yang2025qwen3, wu2025beyond, ma2025slow}. In practice, distillation methods often employ supervised fine-tuning to enable a model to mimic reasoning trajectories generated by large reasoning models~\citep{guo2025deepseek, muennighoffs1, ye2025limo, liu2025qfft}. For instance, S1~\citep{muennighoffs1} and LIMO~\citep{ye2025limo} enhance the reasoning capacity of small language models by fine-tuning models on well-crafted distillation datasets. However, training models on distillation datasets that overlap with benchmark data can inflate the performance of distilled models. Thus, our work aims to develop detection methods for identifying distillation data potentially used in reasoning distillation.

\paragraph{Detecting training data from LLMs.}
Training data detection on LLMs has been studied in previous work, including fine-tuning data detection and pretraining data detection~\citep{mattern2023membership, fu2024membership, zhang2024min, shidetecting, raoof2025infilling}. Training data may pose risks such as privacy leakage, where training data containing personal information may lead to privacy leakage~\citep{grynbaum2023times, mozes2023use}. Additionally, the training dataset may inadvertently include data from benchmarks, which compromises the reliability of benchmark evaluations~\citep{orenproving, choi2025contamination, xu2024benchmarking}. Fine-tuning data detection for LLMs aims to determine the training data used for fine-tuning~\citep{mattern2023membership, mireshghallah2022quantifying}. The repeated exposure of the fine-tuning data across multiple training epochs increases their vulnerability to privacy attacks. Pretraining data detection aims to determine whether a piece of text is included in the pretraining dataset~\citep{zhang2024fine}. The task is particularly challenging due to the massive scale of the pretraining corpus and the fact that pretraining usually runs for only one epoch~\citep{shidetecting}. Previous studies often design scoring functions that compute a score for each input sequence to detect training data from LLMs~\citep{zhang2024min}. This work formulates the problem of distillation data detection and introduces a novel and effective approach that exploits model-generated tokens instead of input sequences.

\paragraph{Distinction from RL post-training data detection} Reinforcement learning post-training data detection aims to detect training data used in the RL post-training phase of large language models~\citep{tao2026detecting, zhang2026detecting}. 
While RL training data detection also focuses on identifying questions used for model training, it differs fundamentally from distillation data detection in the training paradigm, which leads to entirely different signals required for detection. 
RL post-training optimizes via reward maximization rather than likelihood, causing the model to converge to narrow high-reward reasoning paths on seen questions~\citep{tao2026detecting}.
Reasoning distillation is a token-level supervised imitation of teacher-generated CoT trajectories, training models to maximize the likelihood of target responses conditioned on the questions. 
These two paradigms yield fundamentally different signals: entropy collapse~\citep{tao2026detecting} versus token likelihood memorization.
This makes entropy-based methods for RL training data detection inapplicable to distillation data detection, thus necessitating a separate formulation and method.

\section{Experimental Details}
\label{Exp_details}
\subsection{Details of Datasets and Models}
\label{details of dataset}
\paragraph{Datasets and Models.}
To obtain diverse distilled reasoning models, we fine-tune different models on three well-crafted distillation datasets, including S1, S1.1~\citep{muennighoffs1}, and LIMO~\citep{ye2025limo}. Using questions as prompts, developers often generate reasoning trajectories along with the final answer from advanced large reasoning models to construct distillation datasets. The S1 and S1.1 datasets contain 1,000 examples, whose reasoning trajectories are produced from Gemini~\citep{comanici2025gemini} and DeepSeek-R1~\citep{guo2025deepseek}, respectively. LIMO is a high-quality distillation dataset containing only 817 examples, where each example contains a question together with the reasoning trajectory and final answer. We provide details of the distillation datasets in Table~\ref{details_datasets}. As described in Appendix~\ref{sample_ex}, we present the illustration of a sample comprising a question, corresponding reasoning trajectories, and the final answer. In our experiments, following previous work~\citep{muennighoffs1, ye2025limo, liu2025qfft}, we conduct experiments on different base models, including Qwen2.5-Instruct family~\citep{qwen2025qwen25technicalreport}, Llama-3.1-8B-Instruct~\citep{dubey2024llama}, Gemma-7B-it~\citep{team2024gemma}, and Mistral-7B-Instruct~\citep{jiang2023mistral7b} models. Specifically, we fine-tune these models on the distillation dataset, enabling models to imitate reasoning trajectories generated by teacher models.
\begin{table}[ht]
\centering
\footnotesize
\caption{Detailed statistics of distillation datasets.}
\label{details_datasets}
\renewcommand\arraystretch{1.5}
\resizebox{0.9\textwidth}{!}{
\setlength{\tabcolsep}{4mm}{
\begin{tabular}{lccccc}
\toprule
\textbf{Datasets} & \textbf{Samples} & \textbf{Avg. Tokens} & \textbf{Domain} & \textbf{Source model} & \textbf{Link}\\
\hline
LIMO~\citep{ye2025limo}     & 817  & 7126  & Math  & QwQ-32B, etc. & \href{https://huggingface.co/datasets/GAIR/LIMO}{\textcolor{black}{Dataset}}\\
S1~\citep{muennighoffs1}        & 1000 & 5058 & Math, Science & Gemini & \href{https://huggingface.co/datasets/simplescaling/s1K}{\textcolor{black}{Dataset}} \\
S1.1~\citep{muennighoff2025s}        & 1000  &  10038  & Math, Science    & DeepSeek-R1 & \href{https://huggingface.co/datasets/simplescaling/s1K-1.1}{\textcolor{black}{Dataset}} \\

\bottomrule
\end{tabular}

}}
\end{table}

\subsection{Training Details}
\label{training_details}
In our experiments, we fine-tune different base models on distillation datasets to obtain distilled reasoning models. In particular,
we perform full-parameter fine-tuning on 8 A100 GPUs using DeepSpeed ZeRO-3 optimization, with a sequence length limit of 16,384 tokens. In Table~\ref{train_hyperparameters}, we report detailed training parameters employed in the supervised fine-tuning process.
\begin{table}[ht]
\centering
\footnotesize
\caption{Hyperparameters configuration of supervised fine-tuning.}
\label{train_hyperparameters}
\renewcommand\arraystretch{1.2}
\resizebox{0.8\textwidth}{!}{
\setlength{\tabcolsep}{6mm}{
\begin{tabular}{lccc}
\toprule
\textbf{Hyperparameter} & \textbf{S1} & \textbf{S1.1} & \textbf{LIMO} \\
\hline
sequence length        & 16384 & 16384 & 16384 \\
batch size        & 8     & 8    & 8     \\
learning rate     & 1e-5  & 1e-5  & 1e-5  \\
epochs             & 5     & 5     & 5     \\
scheduler      & cosine& cosine& cosine\\
weight decay      & 1e-4  & 1e-4  & 1e-4  \\
zero optimization & ZeRO-3 &ZeRO-3 &ZeRO-3 \\
training precision &bf16 &bf16 &bf16 \\

\bottomrule
\end{tabular}

}}
\end{table}

\subsection{Baseline Methods}
\label{baseline}
We compare our method with current competitive baselines: (1) \textbf{Perplexity}~\citep{li2023estimating}: uses the perplexity of input text as the membership metric. (2) \textbf{Zlib}~\citep{carlini2021extracting}: computes the ratio of example perplexity and zlib compression entropy (3) \textbf{Lowercase}~\citep{carlini2021extracting}: computes the ratio of the perplexity on the text before and after lowercasing. 
(4) \textbf{Neighbor}~\citep{mattern2023membership}: perturbs the input sentence with masked language models to create ``neighbor'' and compares the loss of the input sentence with the average loss of the neighbor sentences.
(5) \textbf{Min-K\%}~\citep{shidetecting}: computes the average log-likelihood of K\% outlier tokens with the smallest predicted probability. 
(6) \textbf{Min-K\%++}~\citep{zhang2024min}: compares the probability of the target token with the expected probability of all tokens within the vocabulary.
(7) \textbf{Infilling Score}~\citep{raoof2025infilling}: computes the ratio of the infilling probability of the ground-truth token and the maximum causal likelihood token. These methods typically use likelihood-based scores derived from input tokens for training data detection, whereas our method leverages output tokens generated by models for detection~\citep{tao2026detecting}.
(8) \textbf{Entropy-Temp} and (9) \textbf{Entropy-Noise} are entropy-based metrics over output tokens.
(10) \textbf{Self-Critique} measures the entropy curve similarity between two sampled reasoning trajectories~\citep{tao2026detecting}. 
Additionally, we introduce two naive variants of our method as ablation baselines, which use generated tokens to determine data membership. Specifically, (11) \textbf{Generated Perplexity}: computes the perplexity using the probabilities of generated tokens. (12) \textbf{Generated Min-K\%}: computes the average log-likelihood of K\% generated tokens with the lowest predicted probability. 

\subsection{Implementation Details and Evaluation Metrics}
\label{implementation}
\paragraph{Implementation details}
To effectively evaluate our method, we fine-tune models (e.g., Qwen2.5-32B-Instruct) separately on S1, S1.1, and LIMO distillation datasets to obtain diverse distilled reasoning models. For the main results, the original datasets are split into training and testing subsets, with an 8:2 train-test split.  We then perform full-parameter fine-tuning on 8 A100 GPUs using DeepSpeed ZeRO-3 optimization, with a sequence length limit of 16,384 tokens. We provide details of model training in the Appendix~\ref{training_details}. To ensure fair evaluation,  we construct balanced datasets of member and non-member samples, drawn respectively from the training and test sets, ensuring an IID setting. In the main experiments, we apply a greedy decoding strategy for generation, and compute the TPD score using the first 300 generated tokens with $\tau = 1$ and $\alpha = 0.6$. For two naive variants of our method, we compute the sample score using only the first 1,000 generated tokens, since using all model outputs incurs prohibitive computational costs due to their excessive length. 

\paragraph{Evaluation metrics}
We evaluate the performance of our method and baselines for detecting distillation data by measuring the following metrics: (1) AUC, the area under the receiver operating characteristic curve; (2) TPR@1\%FPR, the true positive rate at 1\% false positive rate~\citep{carlini2022membership}. Instead of paying equal attention to members and nonmembers, this metric pays more attention to members and evaluates whether one can confidently identify members.

\section{Experimental Results}
\subsection{Challenge of Partial Availability}
\label{evaluation_partial}
Our task of distillation data detection focuses on a realistic setting where only the question of a datapoint is accessible, since accessing question-response pairs is generally infeasible due to the proprietary nature of datasets and the non-deterministic generation process in solution construction. To illustrate the limitations of existing methods due to partial data availability, we conduct experiments on the S1 dataset using Qwen2.5-7B-Instruct under two settings: using only the question and using the question-response pair.

Table~\ref{settings} reports the AUC and TPR@1\%FPR scores of current methods. This validates the unique contribution of this work in the realistic setting. 
When it comes to the settings with access to question-response pairs, all methods can achieve excellent performance in the detection task. We conjecture that the question and response pair contains lots of membership signals that significantly degrade the task difficulty. However, baselines fail to perform detection under the realistic scenario where only the question is accessible at detection time. This finding suggests that the partial availability in distillation data detection undermines the effectiveness of current approaches. 
In contrast, our method achieves robust performance even without relying on model's responses. This demonstrates its unique contribution in a realistic setting, where the auditor has access only to the question.
\begin{table}[!t]
\centering
\footnotesize
\caption{AUC and TPR@1\%FPR scores for our method and baselines on the S1 dataset under two distinct settings: Question-Only (utilizing only the question) and Question-Response (utilizing the question-response pair). \textbf{Bold} indicates the best performance.}
\label{settings}
\renewcommand\arraystretch{1}
\resizebox{0.8\textwidth}{!}{
\setlength{\tabcolsep}{0.8mm}{
\begin{tabular}{lcc|cc}
\toprule
\multirow{2}{*}{\textbf{Method}}& \multicolumn{2}{c}{AUC} & \multicolumn{2}{c}{TPR@1\%FPR} \\
\cmidrule(lr){2-3} \cmidrule(lr){4-5} 
 & \multicolumn{1}{c}{Question-Only} & \multicolumn{1}{c}{Question-Response} & \multicolumn{1}{c}{Question-Only} & \multicolumn{1}{c}{Question-Response} \\
\midrule

Perplexity &0.444	&0.972 &0.040 & 0.745 \\
Lowercase &0.435	&\textbf{0.998} &0.020 & \textbf{0.970} \\ 
Zlib &0.474	&0.940 &0.025 & 0.885 \\
\textsc{Min-k\%} &0.443	&0.972 &0.040 & 0.745 \\	
\textsc{Min-k\%++} &0.472	&0.704 &0.040 & 0.125 \\

\cmidrule(lr){1-5} 
\rowcolor{gray!10} Ours &\textbf{0.855}	& 0.872	&\textbf{0.345} & 0.425\\

\bottomrule
\end{tabular}
}}
\end{table}

\subsection{Token-wise Probability Distribution}
\label{length_compare}
\begin{figure*}[th]
    \centering
    \includegraphics[width=1\textwidth, height=0.32\textheight, keepaspectratio=False]{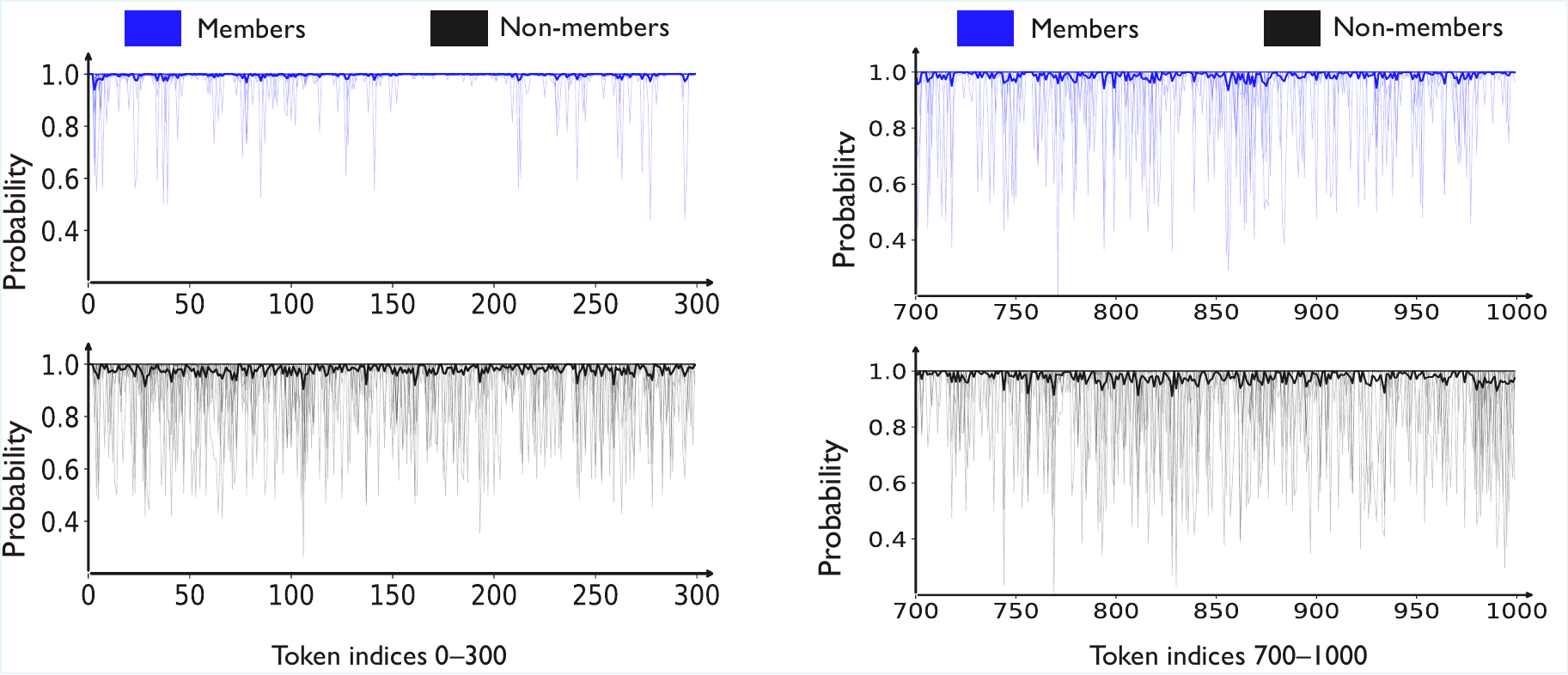}
    \caption{Comparison of token-level generation behaviour of distilled models for 20 members and 20 non-members questions under greedy decoding. \textbf{(Left) Token-wise probability distributions of the first 300 generated tokens}: we contrast the distribution of token-wise probability between members and non-members over the first 300 generated tokens. \textbf{(Right) Token-wise probability distributions of generated tokens at indices 700 to 1000}: we compare the distribution of token-wise probability between members and non-members over generated tokens at indices 700 to 1000.}
    \label{Token-wise}
\end{figure*}

\paragraph{Early generated tokens exhibit higher determinism}
To investigate differences in probability distributions of generated tokens across positions, we analyze tokens at the first 300 positions and at positions 700–1000 for both members and non-members. Figure~\ref{Token-wise} shows the token-wise probability distributions from the distilled reasoning model across 20 member and 20 non-member samples. We observe that the generated tokens at positions 700–1000 exhibit only minor probability distribution differences between members and non-members (Right). This pattern is significantly different from the generation behavior in the first 300 generated tokens (Left), where members tend to produce high-probability tokens, whereas non-members induce more low-probability tokens. The disparity indicates that earlier generated tokens are likely to exhibit distinct membership signals for members and non-members. The near-deterministic behavior in early tokens likely stems from the distillation loss, which trains the model to imitate target reasoning trajectories, producing strongly question-conditioned outputs at the start of generation. As generation proceeds, later generated tokens become increasingly conditioned on the generated context, weakening the near-deterministic behavior observed in early generation.

\subsection{Case study: Detecting Distillation Data with Our TPD.}
\label{case_study}
At detection time, our method requires a threshold to determine whether a given question was used for reasoning distillation. In the experiment, we determine the optimal classification threshold with our TPD by maximizing detection accuracy on the validation set. Specifically, we construct a balanced validation set by randomly sampling 50 members and 50 non-members from the S1 dataset, which is split into training and testing subsets with an 8:2 ratio as described in Section~\ref{exp:setup}, and randomly select 150 members and 150 non-members as the test set.
\begin{table}[!t]
\centering
\footnotesize
\caption{Accuracy, TPR, and FPR scores of our method and baselines for detecting distillation data.}
\label{tab:acc}
\renewcommand\arraystretch{1}
\resizebox{\textwidth}{!}{
\setlength{\tabcolsep}{10mm}{
\begin{tabular}{lc c c}
\toprule
\textbf{Method} & Accuracy $\uparrow$ & TPR $\uparrow$ & FPR $\downarrow$ \\
\midrule
Perplexity~\citep{li2023estimating}  &0.490	&0.860	&0.880\\
Lowercase~\citep{carlini2021extracting} &0.470	&0.920	&0.980\\
Zlib~\citep{carlini2021extracting} &0.440	&0.100	&0.220\\
Neighbor~\citep{mattern2023membership}& 0.517	&0.960	&0.927 \\
\textsc{MIN-K\%}~\citep{shidetecting}  &0.490	&0.820	&0.840\\
\textsc{MIN-K\%++}~\citep{zhang2024min} &0.450	&0.760	&0.860\\
Infilling Score~\citep{raoof2025infilling} &0.543	&0.513	&0.427\\
\cmidrule(lr){1-4} 
\rowcolor{gray!10} Ours &\textbf{0.860}	&0.815	&0.095\\
\bottomrule
\end{tabular}
}}
\end{table}

Table~\ref{tab:acc} shows the accuracy, TPR, and FPR scores of our method and baselines under Qwen2.5-32B-Instruct. The results demonstrate the superior performance of our TPD, which consistently outperforms existing baselines in identifying distillation data in practical scenarios. For example, our method achieves a high accuracy of 0.860 and a TPR of 0.815, while maintaining a low FPR of 0.095. Empirical results demonstrate the practicality of our method in real-world applications.

\subsection{Computational Overhead of Our Method}
\label{overhead}
Our method computes the TPD using only the first 300 generated tokens with their probabilities under greedy decoding, incurring negligible computational overhead. To rigorously quantify the overhead of our method, we measure the time to process 1,000 samples using vLLM on 4 A100 GPUs, finding that our method achieves \textit{a throughput of approximately 26 samples per second (38.47 seconds for 1,000 samples)}, 
demonstrating that it scales readily to real-world applications without imposing a significant computational burden.

\subsection{How Does the Distillation Data Size Affect Our Method?}
\label{data_size}

\begin{wrapfigure}{r}{0.42\linewidth}
    \vspace{-10pt}
    \centering
    \includegraphics[width=\linewidth]{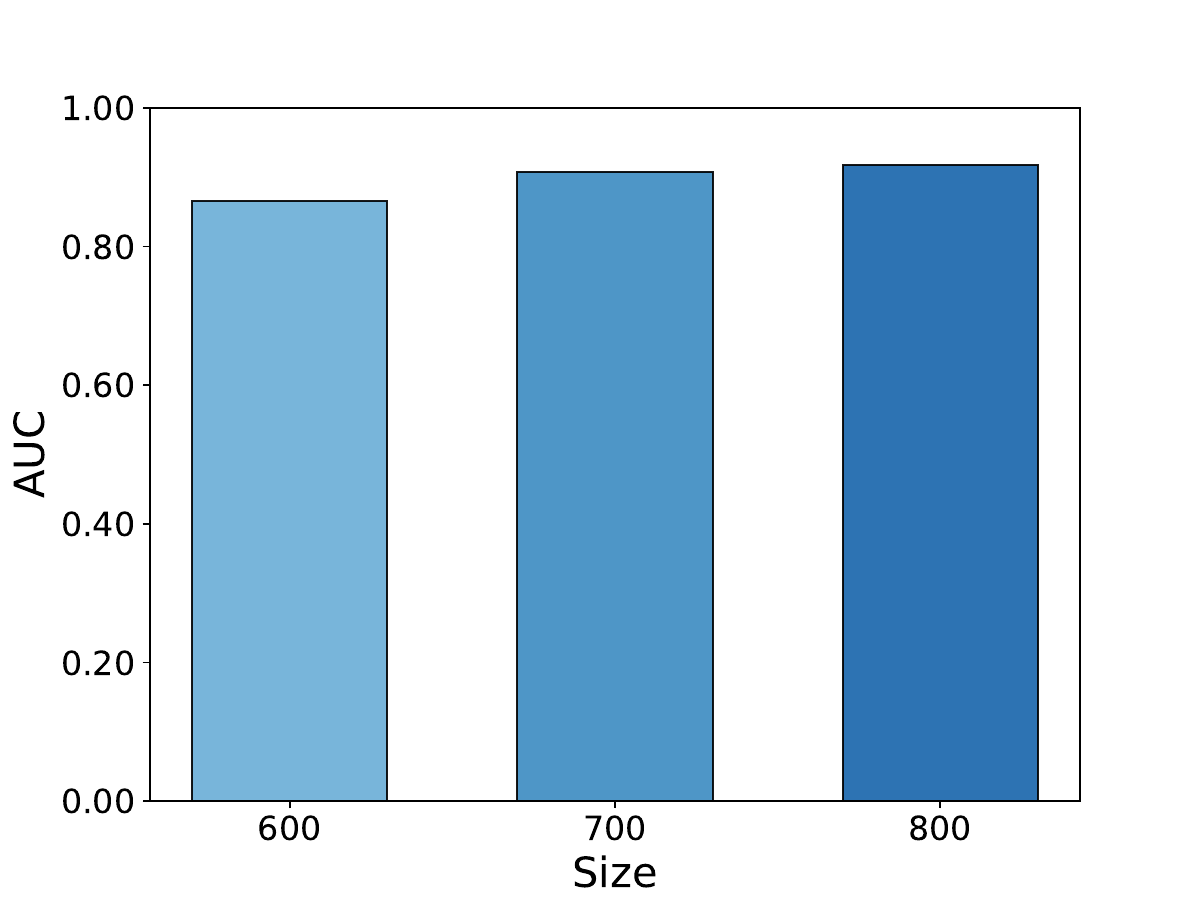}
    \caption{AUC vs. dataset size}
    \label{train_size_auc}
\end{wrapfigure}
To investigate the performance of our method on varying dataset scales, we conduct experiments on the S1 dataset and fine-tune Qwen2.5-32B-Instruct with data sizes of 600, 700, and 800. 
At test time, we construct a balanced dataset to evaluate the performance of our method. 
Figure~\ref{train_size_auc} shows the AUC of our method with various sizes of distillation datasets. The results demonstrate our method consistently achieves reliable detection performance across diverse distilled models.
In addition, we observe that the AUC of our method slightly increases with the size of the distillation dataset, likely because the distilled model trained on more data exhibits enhanced generation behaviour that improves detection. In summary, our experiment indicates the effectiveness of our method with distillation datasets of different sizes.

\subsection{Experiments on the Large-scale Dataset}
\label{exp_larger}
To evaluate the effectiveness of our method on the large-scale dataset, we conduct new experiments on the large dataset OpenR1-Math-220k with Qwen2.5-7B-Instruct. Specifically, we sample 10,000 examples from the original dataset as the training set, and construct a balanced evaluation set of 500 member and 500 non-member samples. As shown in Table~\ref{large_dataset}, our method consistently outperforms all baselines, demonstrating its effectiveness under distillation with the large-scale dataset.
\begin{table}[h]
\centering
\caption{AUC of our method and baselines on OpenR1-Math-220 under Qwen2.5-7B-Instruct.}
\label{large_dataset}
\renewcommand\arraystretch{1.2}
\resizebox{0.95\textwidth}{!}{
\setlength{\tabcolsep}{1mm}{
\begin{tabular}{lcccccccc}
\toprule
\textbf{Method} & Perplexity & Lowercase & Zlib & MIN-K\% & MIN-K\%++ & Generated Perplexity & Generated Min-K & Ours \\
\midrule
\textbf{AUC} & 0.487 & 0.494 & 0.478 & 0.480 & 0.475 & 0.618 & 0.618 & \textbf{0.665} \\
\bottomrule
\end{tabular}}}
\end{table}


\subsection{Is Our Method Affected by Decoding Sampling Parameters?}
\label{parameters}
Our method is motivated by the observation that, under greedy decoding, distilled models generally exhibit different probability distributions for members and non-members. It is natural to wonder: \textit{how does our method perform when decoding with different sampling parameters?} To investigate the performance of our method with different decoding parameters, we conduct experiments on the S1 dataset using Qwen2.5-7B-Instruct with varying temperature (e.g., 0.3, 0.5, 0.7) and top-k (e.g., 10, 30, 50). In the experiment, we set the top‑p sampling parameter to 0.95. 

For the experiment with varying temperature, we set the top-k parameter to 30. For the experiment with varying top-k, the parameter temperature is fixed at 0.3. 
Table~\ref{tab:decoding} presents the AUC score of our method with different sampling parameters. 
The experimental results show that our method achieves better performance at lower temperatures, while its performance remains stable across different Top‑k values.
Notably, our method achieves superior performance under greedy decoding, which aligns with the design rationale of our approach.

\subsection{Experiments on Different Training Configurations}
\label{training_configuration}
We conduct additional experiments on the S1 dataset under various training configurations. Specifically, following previous work, we fine-tune the Qwen2.5-7B-Instruct model with different hyperparameters, including learning rates (e.g., $1 \times 10^{-6}$, $1 \times 10^{-5}$, and $1 \times 10^{-4}$) and numbers of training epochs (e.g., 3, 4, and 5). Table~\ref{tab:training_setups} reports the AUC scores of the baselines and our method under these different settings. The results show that \textbf{our method consistently outperforms all baselines}, demonstrating its effectiveness across a wide range of training setups. We also observe that the performance of our method generally improves with more training epochs, while a learning rate of $1 \times 10^{-6}$ is too small for the model to converge effectively, leading to suboptimal results. In practice, our method is well-suited to real-world scenarios, where training typically adopts appropriate configurations, such as a learning rate of $1 \times 10^{-5}$ and 5 epochs, as used in our experiments.
\begin{table}[t]
\centering
\caption{AUC scores of different methods under various training configurations on the S1 dataset.}
\label{tab:training_setups}
\renewcommand\arraystretch{1}
\resizebox{0.8\textwidth}{!}{
\setlength{\tabcolsep}{3mm}{
\begin{tabular}{lccc|ccc}
\toprule
\multirow{2}{*}{Method} & \multicolumn{3}{c|}{Epoch} & \multicolumn{3}{c}{Learning Rate} \\
\cmidrule(lr){2-4} \cmidrule(lr){5-7}
& 3 & 4 & 5 & $1 \times 10^{-6}$ & $1 \times 10^{-5}$ & $1 \times 10^{-4}$ \\
\midrule
Perplexity & 0.448 & 0.447 & 0.444 & 0.453 & 0.444 & 0.546 \\
Lowercase  & 0.435 & 0.434 & 0.435 & 0.432 & 0.435 & 0.524 \\
Zlib       & 0.477 & 0.476 & 0.474 & 0.477 & 0.474 & 0.549 \\
MIN-K\%    & 0.448 & 0.445 & 0.443 & 0.453 & 0.443 & 0.572 \\
MIN-K\%\texttt{++} & 0.474 & 0.476 & 0.472 & 0.473 & 0.472 & 0.551 \\
Ours        & \textbf{0.579} & \textbf{0.753} & \textbf{0.855} & \textbf{0.495} & \textbf{0.855} & \textbf{0.956} \\
\bottomrule
\end{tabular}
}}
\end{table}


\begin{table}[!t]
\centering
\footnotesize
\caption{AUC scores of our method under varying sampling parameters. Bold shows the superior result.}
\label{tab:decoding}
\renewcommand\arraystretch{1}
\resizebox{0.8\textwidth}{!}{
\setlength{\tabcolsep}{2mm}{
\begin{tabular}{lccccccc}
\toprule
\multirow{2}{*}{Parameters} 
& \multicolumn{3}{c}{Temperature} 
& \multicolumn{3}{c}{Top-K} 
& \multicolumn{1}{c}{Greedy decoding} \\
\cmidrule(lr){2-4} \cmidrule(lr){5-7} \cmidrule(lr){8-8}
& 0.3 & 0.5 & 0.7 & 10 & 30 & 50 & $\backslash$ \\
\midrule
TPD
& 0.712 & 0.679 & 0.689 
& 0.712 & 0.712 & 0.712 
& \textbf{0.855} \\
\bottomrule
\end{tabular}
}}
\end{table}

\subsection{Effectiveness of Our Method under Different Training Data Settings}
\label{scope}
We investigate the scope of our method across three training settings: pretraining, instruction tuning, and reasoning distillation. 
For pretraining data detection, following well-established studies~\citep{maini2024llm, zhang2024fine}, we perform experiments on the Pythia-6.9B model trained on the Pile dataset.
For instruction tuning, we fine-tune Qwen2.5-7B-Base on LIMA~\citep{zhou2023lima}. 
The results in Table~\ref{tab:settings} show that our method underperforms on pretraining data detection, as our method is not tailored for this setting, where detection typically relies on sufficient sample information.
Our method achieves competitive performance in the instruction-tuning setting and yields superior performance on distillation data detection.
These findings suggest that our method can generalize to the instruction-tuning setting, while being particularly effective for distillation data detection. 
We emphasize that our work introduces a new and important task of distillation data detection, where existing methods fail to provide effective detection.
Our proposed method achieves strong performance for detecting distillation data.
\begin{table}[t]
\caption{AUC scores of our method and baselines on the S1 dataset across three training data detection tasks, including pretraining, instruction-tuning, and reasoning distillation. Bold shows the superior result.}
\label{tab:settings}
\centering
\renewcommand\arraystretch{1}
\resizebox{0.8\textwidth}{!}{
\setlength{\tabcolsep}{5mm}{
\begin{tabular}{lccc}
\toprule
\textbf{Method} & \textbf{Pretraining} & \textbf{Instruction-tuning} & \textbf{Distillation} \\
\midrule
Perplexity  & 0.575          & 0.4801          & 0.444          \\
Lowercase   & 0.624          & 0.4666          & 0.435          \\
Zlib        & \textbf{0.676} & 0.4494          & 0.474          \\
Min-k\%     & 0.613          & 0.4800          & 0.443          \\
Min-k\%++   & 0.599          & 0.4909          & 0.472          \\
Ours         & 0.422          & \textbf{0.6364} & \textbf{0.855} \\
\bottomrule
\end{tabular}
}}
\end{table}

\section{Additional Results}
\label{result}
Here, we report the additional experimental results. These results are consistent with the conclusions drawn in the main text. 
Table~\ref{main1_tpr} reports the TPR@1\%FPR of our method and baselines on diverse distilled reasoning models. The results show that our method is effective on various datasets across three different models.
Table~\ref{various_model} reports the AUC score of our method and baselines across diverse models. The results show that our method significantly outperforms the baselines, demonstrating that our method is model-agnostic and broadly applicable for detecting distillation data.
Table~\ref{ai_gen} shows that existing methods on machine text detection fail to detect distillation data, whereas our method achieves superior performance. The empirical results indicate that existing methods for machine-generated text detection fail to generalize to distillation data detection.
Table~\ref{paraphrase} reports AUC scores of our method and baselines under the question paraphrasing setting. 
The results show that our method consistently outperforms baselines, indicating its capability to detect distillation data in the question paraphrasing scenario.
\begin{table*}[th]
\centering
\footnotesize
\caption{TPR@1\%FPR of our method and baselines on diverse distilled models. These models are produced through fine-tuning different-sized models (e.g., Qwen2.5-32B-Instruct) on various distillation datasets, including S1, LIMO, and S1.1 datasets. \textbf{Bold} shows the superior result.}
\label{main1_tpr}
\renewcommand\arraystretch{1}
\resizebox{1.0\textwidth}{!}{
\setlength{\tabcolsep}{1mm}{
\begin{tabular}{lccc|ccc|ccc}
\toprule
\multirow{2}{*}{\textbf{Method}}& \multicolumn{3}{c}{Qwen2.5-7B-Instruct} & \multicolumn{3}{c}{Qwen2.5-14B-Instruct} &\multicolumn{3}{c}{Qwen2.5-32B-Instruct}\\
\cmidrule(lr){2-4} \cmidrule(lr){5-7} \cmidrule(lr){8-10} 
 & \multicolumn{1}{c}{S1} & \multicolumn{1}{c}{LIMO} & \multicolumn{1}{c}{S1.1} & \multicolumn{1}{c}{S1} & \multicolumn{1}{c}{LIMO} & \multicolumn{1}{c}{S1.1} & \multicolumn{1}{c}{S1} & \multicolumn{1}{c}{LIMO} & \multicolumn{1}{c}{S1.1}\\
\midrule
\textit{Input-token-based methods} \\
Perplexity~\citep{li2023estimating} &0.040	&0.000	&0.005 &0.015	&0.012	&0.005 &0.015	&0.006	&0.000 \\
Lowercase~\citep{carlini2021extracting} &0.020	&0.006	&0.010 &0.015	&0.037	&0.000 &0.000	&0.000	&0.000\\
Zlib~\citep{carlini2021extracting} &0.025	&0.000	&0.000 &0.015	&0.006	&0.000 &0.005	&0.000	&0.000\\
Neighbor~\citep{mattern2023membership} &0.025	&0.018	&0.005	&0.030	&0.006	&0.005	&0.025	&0.012	&0.005\\
\textsc{MIN-K\%}~\citep{shidetecting} &0.040	&0.000	&0.005 &0.010	&0.012	&0.010 &0.015	&0.006	&0.000\\
\textsc{MIN-K\%++}~\citep{zhang2024min} &0.040	&0.018	&0.025 &0.000	&0.024	&0.010 &0.025	&0.006	&0.000\\
Infilling Score~\citep{raoof2025infilling} &0.010	&0.006	&0.015	&0.020	&0.000	&0.045	&0.025	&0.018	&0.000\\
\midrule

\textit{Output-token-based methods}\\
Entropy-Temp~\citep{tao2026detecting} &0.060	&0.012	&0.000	&0.125	&0.012	&0.005	&0.130	&0.030	&0.005\\
Entropy-Noise~\citep{tao2026detecting} &0.030	&0.012	&0.025	&0.010	&0.037	&0.020	&0.025	&0.012	&0.040\\
Self-Critique~\citep{tao2026detecting} &0.015	&0.012	&0.005	&0.005	&0.018	&0.000	&0.005	&0.030	&0.000\\
Generated Perplexity &0.235	&0.128	&0.070 &0.350	&0.171	&0.045  &0.160	&0.226	&0.080\\
Generated MIN-K\% &0.235	&0.128	&0.070 &0.350	&0.171	&0.045 &0.160	&0.226	&0.080\\
\rowcolor{gray!10} 
Ours &\textbf{0.345}	&\textbf{0.256}	&\textbf{0.095} &\textbf{0.375}	&\textbf{0.226}	&\textbf{0.090} &\textbf{0.470}	&\textbf{0.335}	&\textbf{0.110}\\

\bottomrule
\end{tabular}
}}
\end{table*}

\begin{table}[th]
\centering
\footnotesize
\caption{AUC scores of our method and baselines on the S1 dataset across various models, including Llama-3.1-8B, Gemma-7b and Mistral-7B. \textbf{Bold} shows the superior result.}
\label{various_model}
\renewcommand\arraystretch{1}
\resizebox{0.8\textwidth}{!}{
\setlength{\tabcolsep}{4mm}{
\begin{tabular}{lccc}
\toprule
\multirow{2}{*}{\textbf{Method}}& \multicolumn{3}{c}{AUC}  \\
\cmidrule(lr){2-4} 
 & \multicolumn{1}{c}{Llama-3.1-8B} & \multicolumn{1}{c}{Gemma-7b} & \multicolumn{1}{c}{Mistral-7B} \\
\midrule

Perplexity~\citep{li2023estimating}& 0.529	&0.537	& 0.549\\
Lowercase~\citep{li2023estimating} & 0.524	&0.537	& 0.486 \\
Zlib~\citep{carlini2021extracting} & 0.539	&0.533	& 0.547	\\
Neighbor~\citep{mattern2023membership} & 0.522 &0.554	 &0.522 \\
\textsc{Min-k\%}~\citep{shidetecting}  & 0.554	&0.535	& 0.564	\\
\textsc{Min-k\%++}~\citep{zhang2024min} & 0.562	&0.532	& 0.543\\
Infilling Score~\citep{raoof2025infilling} &0.557 &0.522 &0.550 \\
\cmidrule(lr){1-4} 
\rowcolor{gray!10} Ours & \textbf{0.927} &\textbf{0.943} &\textbf{0.953} \\

\bottomrule
\end{tabular}
}}
\end{table}

\begin{table}[!t]
\centering
\footnotesize
\caption{Comparison of AUC scores between our method and baselines under question paraphrasing, evaluated across different models. \textbf{Bold} shows the superior result.}
\label{paraphrase}
\renewcommand\arraystretch{1}
\resizebox{0.8\textwidth}{!}{
\setlength{\tabcolsep}{3mm}{
\begin{tabular}{lccc}
\toprule
\multirow{1}{*}{\textbf{Method}}& \multicolumn{1}{c}{Qwen2.5-7B} & \multicolumn{1}{c}{Qwen2.5-14B} & \multicolumn{1}{c}{Qwen2.5-32B} \\
\midrule
Perplexity~\citep{li2023estimating} &0.463	&0.468	&0.469\\
Lowercase~\citep{carlini2021extracting} &0.503	&0.493	&0.543\\
Zlib~\citep{carlini2021extracting} &0.497	&0.497	&0.496\\
Neighbor~\citep{mattern2023membership} &0.538	&0.545	&0.557 \\
\textsc{Min-k\%}~\citep{shidetecting} &0.469	&0.475	&0.473\\
\textsc{Min-k\%++}~\citep{zhang2024min} &0.500	&0.494	&0.527\\
Infilling Score~\citep{raoof2025infilling} &0.529	&0.534	&0.574\\
\cmidrule(lr){1-4} 
\rowcolor{gray!10} Ours &\textbf{0.615}	&\textbf{0.692}	&\textbf{0.691} \\
\bottomrule
\end{tabular}
}}
\end{table}

\begin{table}[!t]
\centering
\footnotesize
\caption{AUC score of our method and baselines of machine text detection on S1 dataset, evaluated across different-sized models. \textbf{Bold} shows the superior result.}
\label{ai_gen}
\renewcommand\arraystretch{1}
\resizebox{0.8\textwidth}{!}{
\setlength{\tabcolsep}{3mm}{
\begin{tabular}{lccc}
\toprule
\multirow{1}{*}{\textbf{Method}}& \multicolumn{1}{c}{Qwen2.5-7B} & \multicolumn{1}{c}{Qwen2.5-14B} & \multicolumn{1}{c}{Qwen2.5-32B} \\
\midrule
DetectGPT~\citep{mitchell2023detectgpt} &0.495	&0.512	&0.501\\
Fast-DetectGPT~\citep{bao2024fastdetectgpt} &0.443	&0.459	&0.437\\
Binoculars\citep{hans2024spotting} &0.480	&0.481	&0.442\\
DetectLLM~\citep{su2023detectllm} &0.430	&0.441	&0.435\\
\cmidrule(lr){1-4} 
\rowcolor{gray!10} Ours &\textbf{0.855}	&\textbf{0.870}	&\textbf{0.918} \\

\bottomrule
\end{tabular}
}}
\vspace{-4mm}
\end{table}

\section{Implementation Examples}
\label{sample_ex}
We provide examples of distillation data and question prompts in Example~\ref{example_train} and Example~\ref{example_prompt}. A datapoint from the S1 distillation dataset~\citep{muennighoffs1} is presented in Example D.1, comprising a question, corresponding reasoning trajectories, and final answer. In Example D.2, we present three question prompts using the system template.
\begin{promptbox}[example_train]{Illustrative case of distillation data}
\textbf{\textcolor{red}{Question:}}\\
Given a rational number, write it as a fraction in lowest terms and calculate the product of the resulting numerator and denominator. For how many rational numbers between 0 and 1 will $20_{}^{}!$ be the resulting product?

\rule{\linewidth}{0.6pt}

\textbf{\textcolor{red}{Thinking Trajectories:}} \\
The problem asks for the number of rational numbers between 0 and 1 such that when the rational number is written as a fraction in lowest terms, the product of the numerator and the denominator is $20!$.\\[0em]

Let the rational number be $r$, where $0 < r < 1$. We can write $r$ as a fraction $\frac{a}{b}$, where $a$ and $b$ are positive integers, and $\gcd(a, b) = 1$. Since $0 < r < 1$, we have $0 < \frac{a}{b} < 1$, which implies $0 < a < b$. The product of the numerator and the denominator of the fraction in lowest terms is $a \times b$. We are given that $a \times b = 20!$. We need to find the number of pairs $(a, b)$ such that: 

1. $a$ and $b$ are positive integers. 

2. $\gcd(a, b) = 1$. 

3. $a < b$. 

4. $a \times b = 20!$.

Let $N = 20!$. We have $a \times b = N$. Since $\gcd(a, b) = 1$, the prime factors of $N$ are distributed between $a$ and $b$ such that no prime factor is shared by both. Consider the prime factorization of $N = 20!$. Let $N = p_1^{e_1} p_2^{e_2} \cdots p_k^{e_k}$, where $p_i$ are distinct prime numbers and $e_i \ge 1$.\\[0em]

Since $a \times b = N$ and $\gcd(a, b) = 1$, each prime power $p_i^{e_i}$ must be entirely a factor of $a$ or entirely a factor of $b$. For each prime power $p_i^{e_i}$, there are two choices: either $p_i^{e_i}$ is a factor of $a$, or $p_i^{e_i}$ is a factor of $b$. Since there are $k$ distinct prime factors, there are $2^k$ ways to distribute the prime power factors between $a$ and $b$. Each way corresponds to a unique pair $(a, b)$ such that $a \times b = N$ and $\gcd(a, b) = 1$. Let the number of distinct prime factors of $N = 20!$ be $\omega(20!)$. Then the number of pairs $(a, b)$ such that $a \times b = 20!$ and $\gcd(a, b) = 1$ is $2^{\omega(20!)}$. Now, we need to consider the condition $a < b$.Since $a \times b = N$, if $a \neq b$, then either $a < b$ or $b < a$. If $a = b$, then $a^2 = N$, which means $N$ is a perfect square. $20!$ is not a perfect square because the exponent of some prime factors in its prime factorization is odd (e.g., the exponent of the largest prime less than or equal to 20, which is 19). The pairs $(a, b)$ such that $a \times b = N$ and $\gcd(a, b) = 1$ can be grouped into pairs of the form $(a, b)$ and $(b, a)$, where $a \neq b$. The number of such pairs is $2^{\omega(N)}$. Each pair contributes to either $a < b$ or $b < a$. Since $a \neq b$, exactly half of these pairs will satisfy $a < b$.\\[0em]

The number of pairs $(a, b)$ such that $a \times b = N$, $\gcd(a, b) = 1$, and $a < b$ is $\frac{1}{2} \times 2^{\omega(N)} = 2^{\omega(N) - 1}$. We need to find the number of distinct prime factors of $20!$. The prime numbers less than or equal to 20 are 2, 3, 5, 7, 11, 13, 17, 19. There are 8 distinct prime factors of $20!$. So, $\omega(20!) = 8$. The number of pairs $(a, b)$ satisfying the conditions is $2^{8 - 1} = 2^7 = 128$. Each such pair $(a, b)$ corresponds to a unique rational number $\frac{a}{b}$ between 0 and 1 in lowest terms, where $a \times b = 20!$. The number of rational numbers between 0 and 1 for which the product of the numerator and denominator of the fraction in lowest terms is $20!$ is $128$. Let's verify the reasoning. Consider a simpler example: Product is 12. Prime factorization of 12 is $2^2 \times 3^1$. Distinct prime factors are 2 and 3. $\omega(12) = 2$. Pairs $(a, b)$ such that $a \times b = 12$ and $\gcd(a, b) = 1$: Factors of 12 are 1, 2, 3, 4, 6, 12. Possible pairs $(a, b)$ with $a \times b = 12$: (1, 12), (2, 6), (3, 4). Check $\gcd(a, b) = 1$: $\gcd(1, 12) = 1$ $\gcd(2, 6) = 2 \neq 1$ $\gcd(3, 4) = 1$ \\ The pairs $(a, b)$ such that $a \times b = 12$ and $\gcd(a, b) = 1$ are formed by distributing the prime factors $2^2$ and $3^1$. $a$ can have factors from $\{1, 2^2, 3^1, 2^2 \times 3^1\}$ and $b$ will have the remaining factors. Case 1: $a = 1, b = 12$. $\gcd(1, 12) = 1$. Case 2: $a = 4, b = 3$. $\gcd(4, 3) = 1$. Case 3: $a = 3, b = 4$. $\gcd(3, 4) = 1$. Case 4: $a = 12, b = 1$. $\gcd(12, 1) = 1$. \\[0em]

Something is wrong here. Let's use the prime factorization directly. $12 = 2^2 \times 3^1$. Prime factors are 2 and 3. Possible distributions of prime powers:- $a$ has $2^0, 3^0$, $b$ has $2^2, 3^1 \implies a = 1, b = 12$. $\gcd(1, 12) = 1$.- $a$ has $2^2, 3^0$, $b$ has $2^0, 3^1 \implies a = 4, b = 3$. $\gcd(4, 3) = 1$.- $a$ has $2^0, 3^1$, $b$ has $2^2, 3^0 \implies a = 3, b = 4$. $\gcd(3, 4) = 1$.- $a$ has $2^2, 3^1$, $b$ has $2^0, 3^0 \implies a = 12, b = 1$. $\gcd(12, 1) = 1$. The pairs $(a, b)$ such that $a \times b = 12$ and $\gcd(a, b) = 1$ are (1, 12) and (4, 3) and (3, 4) and (12, 1). Number of such pairs is $2^{\omega(12)} = 2^2 = 4$. Now, consider the condition $a < b$. Pairs where $a < b$: (1, 12) and (3, 4). Number of such pairs is $\frac{1}{2} \times 4 = 2$. The rational numbers are $\frac{1}{12}$ and $\frac{3}{4}$. Product of numerator and denominator: $1 \times 12 = 12$, $3 \times 4 = 12$. Fractions are in lowest terms. Rational numbers are between 0 and 1. The number of distinct prime factors of $20!$ needs to be correctly calculated. Prime numbers $\le 20$: 2, 3, 5, 7, 11, 13, 17, 19. There are 8 distinct prime factors. Let $N = 20!$. We are looking for rational numbers $\frac{a}{b}$ such that $\gcd(a, b) = 1$, $0 < a < b$, and $a \times b = N$. The number of pairs $(a, b)$ such that $a \times b = N$ and $\gcd(a, b) = 1$ is $2^{\omega(N)}$. Each pair corresponds to a factorization of $N$ into two coprime factors. Let $N = p_1^{e_1} p_2^{e_2} \cdots p_k^{e_k}$.$a = p_1^{\alpha_1} p_2^{\alpha_2} \cdots p_k^{\alpha_k}$$b = p_1^{\beta_1} p_2^{\beta_2} \cdots p_k^{\beta_k}$ \\[0em]

Since $\gcd(a, b) = 1$, for each $i$, either $\alpha_i = e_i, \beta_i = 0$ or $\alpha_i = 0, \beta_i = e_i$. This gives $2^k$ choices for the pair $(a, b)$. We need $a < b$. Consider the case $a = b$. This happens only if all prime exponents $e_i$ are even, which is not the case for $20!$. So, for every pair $(a, b)$ with $a \times b = N$ and $\gcd(a, b) = 1$, we have $a \neq b$. The pairs can be grouped into $(a, b)$ and $(b, a)$, where $a \neq b$. The number of such pairs is $2^{\omega(N)}$. The number of pairs with $a < b$ is $\frac{1}{2} \times 2^{\omega(N)} = 2^{\omega(N) - 1}$. For $N = 20!$, $\omega(N) = 8$. The number of pairs $(a, b)$ is $2^{8 - 1} = 2^7 = 128$. Each pair $(a, b)$ corresponds to a unique rational number $\frac{a}{b}$ between 0 and 1 in lowest terms, such that $a \times b = 20!$. Double-check the problem statement and constraints. Rational numbers between 0 and 1. Resulting product of numerator and denominator is $20!$. Consider a rational number $r$ between 0 and 1. When written as a fraction in lowest terms, let it be $\frac{a}{b}$, where $\gcd(a, b) = 1$ and $0 < a < b$. We are given $a \times b = 20!$. We need to find the number of such fractions $\frac{a}{b}$. This is equivalent to finding the number of pairs $(a, b)$ satisfying the conditions. The number of distinct prime factors of $20!$ is the number of primes less than or equal to 20, which is 8. The primes are 2, 3, 5, 7, 11, 13, 17, 19. Let the prime factorization of $20!$ be $p_1^{e_1} p_2^{e_2} \cdots p_8^{e_8}$.$a \times b = p_1^{e_1} p_2^{e_2} \cdots p_8^{e_8}$. Since $\gcd(a, b) = 1$, for each prime factor $p_i$, either $p_i^{e_i}$ is a factor of $a$ or $p_i^{e_i}$ is a factor of $b$.\\[0em]

Consider the divisors of $20!$ that are coprime to their cofactors in the division. If $d$ is a divisor of $N$, then $N = d \times \frac{N}{d}$. We need $\gcd(d, \frac{N}{d}) = 1$. This means that the prime factors of $d$ and $\frac{N}{d}$ are disjoint. Equivalently, for each prime factor $p_i$ of $N$, either $p_i^{e_i}$ divides $d$ or $p_i^{e_i}$ divides $\frac{N}{d}$. Let $a$ be a divisor of $20!$ such that $\gcd(a, \frac{20!}{a}) = 1$. Then $b = \frac{20!}{a}$. The number of such divisors $a$ is $2^{\omega(20!)}$. These divisors correspond to the possible values of the numerator $a$ in the pairs $(a, b)$ where $a \times b = 20!$ and $\gcd(a, b) = 1$. The pairs are formed by choosing, for each prime power $p_i^{e_i}$, whether it belongs to $a$ or $b$. Example: $12 = 2^2 \times 3^1$. Prime powers are $2^2$ and $3^1$. Choices for factors of $a$: $\{1, 4\}$, choices for factors of $b$: $\{1, 3\}$. Possible values of $a$: $1 \times 1 = 1$, $1 \times 3 = 3$, $4 \times 1 = 4$, $4 \times 3 = 12$. If $a = 1$, $b = 12$. If $a = 3$, $b = 4$. If $a = 4$, $b = 3$. If $a = 12$, $b = 1$. The pairs $(a, b)$ such that $a \times b = N$ and $\gcd(a, b) = 1$ are formed by partitioning the set of prime power factors into two sets, one for $a$ and one for $b$. Let the prime power factors be $P_1, P_2, \ldots, P_k$, where $P_i = p_i^{e_i}$.$a$ is a product of some of these $P_i$, and $b$ is the product of the remaining ones. For each $P_i$, it either goes to $a$ or $b$. $2^k$ possibilities.\\
[0em]

We need $a < b$. Since $a \times b = N$, and $a \neq b$ (as $N$ is not a perfect square), either $a < b$ or $b < a$. The number of pairs with $a < b$ is equal to the number of pairs with $b < a$. Total number of pairs is $2^k$. Number of pairs with $a < b$ is $\frac{2^k}{2} = 2^{k-1}$. Here $k = \omega(20!) = 8$. Number of pairs $(a, b)$ with $a < b$, $\gcd(a, b) = 1$, $a \times b = 20!$ is $2^{8 - 1} = 2^7 = 128$. Each such pair $(a, b)$ corresponds to a unique rational number $\frac{a}{b}$ between 0 and 1 in lowest terms, with the product of numerator and denominator equal to $20!$. Example revisited: Product is 12. Prime factors 2, 3. $\omega(12) = 2$. Pairs $(a, b)$ with $a \times b = 12, \gcd(a, b) = 1$:$a = 1, b = 12$$a = 3, b = 4$$a = 4, b = 3$$a = 12, b = 1$ Pairs with $a < b$: (1, 12), (3, 4). Number = $2^{2 - 1} = 2^1 = 2$. Rational numbers: $\frac{1}{12}, \frac{3}{4}$. Consider the edge cases or potential pitfalls. The condition \"between 0 and 1\" ensures $a < b$. The condition \"fraction in lowest terms\" ensures $\gcd(a, b) = 1$. The problem boils down to finding the number of ways to factor $20!$ into two coprime factors $a$ and $b$ such that $a < b$. Final confirmation of the number of distinct prime factors of $20!$: Primes $\le 20$: 2, 3, 5, 7, 11, 13, 17, 19. There are 8 distinct prime factors. The number of rational numbers is the number of pairs $(a, b)$ satisfying the conditions, which is $2^{\omega(20!) - 1} = 2^{8 - 1} = 2^7 = 128$. The question asks for \"how many rational numbers\", so the answer should be the number of such fractions. Each valid pair $(a, b)$ corresponds to one rational number. If the product was a perfect square, say 36.$36 = 2^2 \times 3^2$. $\omega(36) = 2$. Pairs $(a, b)$ with $a \times b = 36, \gcd(a, b) = 1$:\$a=1, b=36$\$a=4, b=9$ $a=9, b=4$ $a=36, b=1$ Here $a \neq b$. If the product was $p^e$, where $p$ is prime. Pairs $(1, p^e)$ and $(p^e, 1)$. If $e > 0$, $1 < p^e$. Rational number $\frac{1}{p^e}$.  If the product was $pq$, where $p, q$ are distinct primes. Pairs $(1, pq), (p, q), (q, p), (pq, 1)$. If $p < q$, then $p < q$, $1 < pq$. Pairs with $a < b$: $(1, pq), (p, q)$. Number = $2^{2 - 1} = 2$. Rational numbers: $\frac{1}{pq}, \frac{p}{q}$. The reasoning seems consistent. The number of rational numbers is indeed $2^{\omega(20!) - 1}$.

\rule{\linewidth}{0.6pt}

\textbf{\textcolor{red}{Answer: }} \\
Solution: Let the rational number between 0 and 1 be $r$. When written as a fraction in lowest terms, let it be $\frac{a}{b}$, where $a$ and $b$ are positive integers, $\gcd(a, b) = 1$, and $0 < a < b$.
We are given that the product of the numerator and denominator is $20!$, so $a \times b = 20!$.
We need to find the number of pairs $(a, b)$ such that $a \times b = 20!$, $\gcd(a, b) = 1$, and $a < b$. Let $N = 20!$. The prime factorization of $N$ is $p_1^{e_1} p_2^{e_2} \cdots p_k^{e_k}$, where $p_i$ are distinct prime numbers. The number of distinct prime factors of $20!$ is the number of primes less than or equal to 20. These primes are 2, 3, 5, 7, 11, 13, 17, 19. Thus, there are 8 distinct prime factors, so $k = 8$.\\[0em]

Since $a \times b = N$ and $\gcd(a, b) = 1$, each prime power factor $p_i^{e_i}$ of $N$ must be entirely a factor of $a$ or entirely a factor of $b$.
The number of ways to distribute the prime power factors between $a$ and $b$ is $2^k$, where $k$ is the number of distinct prime factors of $N$. Each distribution corresponds to a unique pair $(a, b)$ such that $a \times b = N$ and $\gcd(a, b) = 1$.
The number of such pairs is $2^{\omega(N)}$, where $\omega(N)$ is the number of distinct prime factors of $N$.\\[0em]

Since $20!$ is not a perfect square, for every pair $(a, b)$ such that $a \times b = 20!$ and $\gcd(a, b) = 1$, we have $a \neq b$. The pairs can be grouped into $(a, b)$ and $(b, a)$, where $a \neq b$.
The number of pairs with $a < b$ is equal to the number of pairs with $b < a$.
The total number of pairs is $2^{\omega(20!)}$. The number of pairs with $a < b$ is $\frac{1}{2} \times 2^{\omega(20!)} = 2^{\omega(20!) - 1}$.\\[0em]

The number of distinct prime factors of $20!$ is $\omega(20!) = 8$.
The number of pairs $(a, b)$ such that $a \times b = 20!$, $\gcd(a, b) = 1$, and $a < b$ is $2^{8 - 1} = 2^7 = 128$. Each such pair $(a, b)$ corresponds to a unique rational number $\frac{a}{b}$ between 0 and 1 in lowest terms, with the product of the numerator and denominator equal to $20!$.\\[0em]

Final Answer: The final answer is $\boxed{128}$
\end{promptbox}

\vspace{10mm}
\begin{promptbox}[example_prompt]{Illustrative case of prompts}
\textbf{\textcolor{blue}{Prompt1:}}
{\fontsize{9pt}{9pt}\selectfont
\begin{verbatim}
<|im_start|>system
You are Qwen, created by Alibaba Cloud. You are a helpful assistant.<|im_end|>
<|im_start|>user 
Given a rational number, write it as a fraction in lowest terms and calculate 
the product of the resulting numerator and denominator. For how many rational 
numbers between 0 and 1 will $20_{}^{}!$ be the resulting product? <|im_end|>
<|im_start|>assistant 
\end{verbatim}
}
\rule{\linewidth}{0.6pt}

\textbf{\textcolor{blue}{Prompt2:}}
{\fontsize{9pt}{9pt}\selectfont
\begin{verbatim}
<|im_start|>system
You are Qwen, created by Alibaba Cloud. You are a helpful assistant.<|im_end|>
<|im_start|>user
Joanie takes a $\$6,\!000$ loan to pay for her car. The annual interest rate on 
the loan is $12\%$. She makes no payments for 4 years, but has to pay back all 
the money she owes at the end of 4 years. How much more money will she owe if 
the interest compounds quarterly than if the interest compounds annually? 
Express your answer as a dollar value to the nearest cent.<|im_end|>
<|im_start|>assistant
\end{verbatim}
}
\rule{\linewidth}{0.6pt}
\textbf{\textcolor{blue}{Prompt3:}}
{\fontsize{9pt}{9pt}\selectfont
\begin{verbatim}
<|im_start|>system
You are Qwen, created by Alibaba Cloud. You are a helpful assistant.<|im_end|>
<|im_start|>user
Gary purchased a large beverage, but only drank $m/n$ of it, where $m$ and $n$ 
are relatively prime positive integers. If he had purchased half as much and 
drunk twice as much, he would have wasted only $2/9$ as much beverage. 
Find $m+n$.<|im_end|>
<|im_start|>assistant
\end{verbatim}
}

\end{promptbox}


\end{document}